%% file: main.tex
\documentclass[runningheads]{llncs}

\usepackage[final,year=2026,ID=4121]{eccv}
\usepackage{eccvabbrv}

\usepackage{graphicx}
\usepackage{booktabs}
\usepackage{multirow}
\usepackage{diagbox}
\usepackage{pifont}
\usepackage{colortbl}
\usepackage[american]{babel}
\usepackage{booktabs,makecell}
\usepackage{bm}
\usepackage{enumitem}
\usepackage{cuted}
\usepackage{caption}
\usepackage{subcaption}
\usepackage{wrapfig}
\usepackage[accsupp]{axessibility}  % Improves PDF readability for those with disabilities.
\definecolor{mygreen}{RGB}{0,150,0}
\definecolor{NiceRed}{RGB}{204, 51, 51}

\usepackage{hyperref}

\usepackage{orcidlink}

\begin{document}

\title{FitControler: Toward Fit-Aware Virtual Try-On} 


\author{Lu Yang\inst{1} \and
Yicheng Liu\inst{1} \and
Letian Zhou\inst{1} \and
Yanan Li \inst{2} \and
Xiang Bai \inst{1} \and
Hao Lu \inst{1}$^,$\thanks{Corresponding author.}
% $^,$\textsuperscript{\ding{41}}
}

\authorrunning{L.Yang~ et al.}

\institute{Huazhong University of Science and Technology, China \and
Wuhan Institute of Technology, China\\
\email{\{lu\_yang1, hlu\}@hust.edu.cn}
}

\maketitle

\vspace{-25pt}
\begin{figure}[ht]
    \small\centering
    \includegraphics[width=\textwidth]{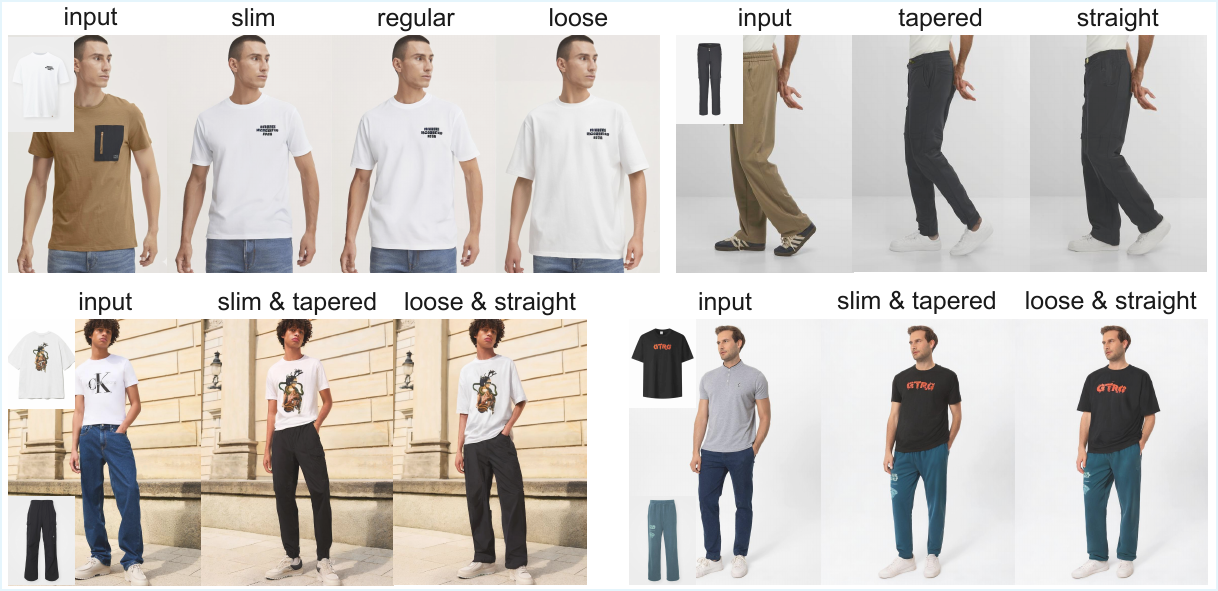}
    \vspace{-20pt}
    \caption{\textbf{Illustrations of FitControler generations}. % We present FitControler, a plug-in module for diffusion-based virtual try-on models that enable customized control of the garment fit.
    Zoom in and compare contour differences of the garment fit.}
    \label{fig:abs}
\end{figure}
\vspace{-30pt}

\begin{abstract}
Realistic virtual try-on (VTON) concerns not only faithful rendering of garment details but also coordination of the style. 
Prior art typically pursues the former, but neglects a key factor that shapes the holistic style---garment fit.
Garment fit delineates how a garment aligns with the body of a wearer and is a fundamental element in fashion design. 
In this work, we introduce fit-aware VTON and present FitControler, a learnable plug-in that can seamlessly integrate into modern VTON models to enable customized fit control. 
To achieve this, we highlight two challenges: i) how to delineate layouts of different fits and ii) how to render the garment that matches the layout.
FitControler first features a fit-aware layout generator to redraw the body-garment layout conditioned on a set of delicately processed garment-agnostic representations, and a multi-scale fit injector is then used to deliver layout cues to enable layout-driven VTON.
In particular, we build a fit-aware VTON dataset termed Fit4Men, including $13,000$ body-garment pairs of different fits, covering both tops and bottoms, and featuring varying camera distances and body poses. Two fit consistency metrics are also introduced to assess the fitness of generations.
Extensive experiments show that FitControler can work with various VTON models and achieve accurate fit control. 
Code and data are released 
at \url{https://github.com/tiny-smart/FitControler}.
\end{abstract}

% \newpage

%%%%%%%%% BODY TEXT
\section{Introduction}
\label{sec:intro}

\begin{wrapfigure}{r}{0.5\linewidth}
\centering  
\begin{center}
\vspace{-30pt}
\includegraphics[width=\linewidth]{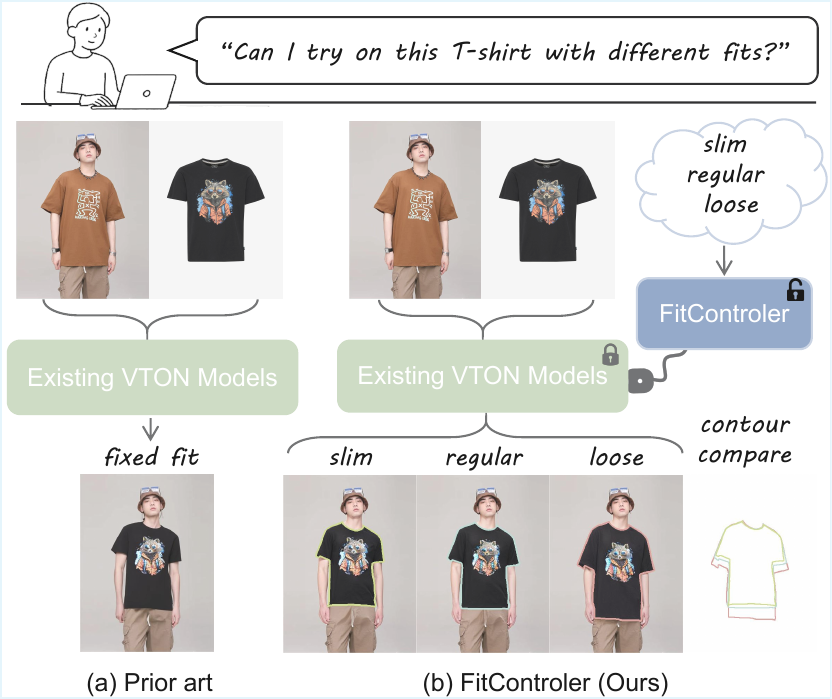}
\end{center}
\vspace{-10pt}
\caption{\textbf{Comparison between existing VTON models and our proposed FitControler.} (a) Existing models produce only a fixed fit, often leading to unnatural results due to mismatched fit. (b) With FitControler, the same inputs can produce try-on results with customized fits such as \emph{slim}, \emph{regular}, and \emph{loose} T-shirts, which much better matches the overall style and user preference.}
\label{fig:intro}  
\vspace{-5pt}
\end{wrapfigure}

\emph{``Does this outfit suit me?"}---this is a common question for online shoppers, yet difficult to answer through imagination alone. VTON seeks to bridge this gap by generating realistic try-on images of a wearer with desired garments. 
Realistic VTON generation, however, is difficult. It hinges on not only faithful preservation of garment appearance and body pose but also coordination of the holistic style.  
Existing work~\cite{choi2021viton,kim2024stableviton,choi2024improving,chong2024catvton,zhou2025learning,wan2025incorporating} mainly focuses on recovering garment details while overlooking the style consistency. 
Per Fig.~\ref{fig:intro}(a), this negligence can result in disharmony of the feeling and may mismatch user preferences.

In the fashion domain, style is primarily influenced by three factors: color coordination, ways of wearing, and fit~\cite{kollnitz2021fashion,fan2004clothing}. Since the garment is preset, the latter two matter in VTON. 
Ways of wearing---such as tucking in the hem or rolling up sleeves---affect local styles, while \emph{fit shapes the holistic style}, delineates how a garment aligns with the body, and is governed by physical factors such as garment proportions, measurements, and ease allowance.
These physical properties manifest visually via variations in tightness (\eg, slim vs.\ loose T-shirts) and shape (\eg, tapered vs.\ straight trousers)~\cite{fan2004clothing}, making it possible to reproduce target fits visually without the need for precise physical modeling.
% Some works~\cite{yan2023linking,chen2024wear} have enabled control over the ways of wearing, but fit remains underexplored. 

In this work, we introduce fit-aware VTON and present FitControler, a novel plug-in engineered for seamless integration with modern diffusion-based VTON models, empowering users with customized fit control (Fig.~\ref{fig:intro}(b)).
We remark that a key challenge for fit-aware VTON lies in \textit{how to ground the abstract notion of fit into tangible visual representations}.
To this end, FitControler follows a two-stage design: i)
delineating the spatial layout between the garment and body given a certain fit, and ii) rendering the garment onto the body conditioned on the layout.

The first stage deals with spatial modeling between the body and garment. Following~\cite{yu2019vtnfp,choi2021viton}, we use the segmentation map to represent the body-garment layout and introduce a fit-aware layout generator to redraw the target segmentation map conditioned on the fit label. 
During layout generation, a phenomenon called \textit{garment shape leakage} arises. 
This leads to a problem where the generator is biased toward replicating the original garment shape, instead of generating a new one.
To address this, we redesign the garment-agnostic representation preprocessor to produce a rectangular mask and a dense pose conditioned on body proportions.
The second stage follows a conditional image generation pipeline.
Unlike previous approaches~\cite{zhang2023adding,mou2024t2i} that often rely on an extra encoder to extract conditional image features, we repurpose the features from the layout generator, and a lightweight multi-scale fit injector is used to deliver the layout features to off-the-shelf VTON models via the ControlNet~\cite{zhang2023adding} interface. 
We find that such a simplification not only reduces the computational overhead but also accelerates training convergence.

In particular, we harvest a fit-orientated VTON dataset, termed Fit4Men. We focus on men garments because they feature limited but well-defined garment styles compared with more varied women garments, making them suitable for an initial exploration on this topic.
Fit4Men features $5,000$ pairs of men T-shirts with $\tt slim$, $\tt regular$, and $\tt loose$ fits and $8,000$ pairs of men trousers with $\tt tapered$ and $\tt straight$ fits. The dataset covers diverse body poses and varying camera distances. 
To objectively assess the fitness of try-on generations, % after VTON, 
two fit consistency metrics measuring % silhouette
contour differences are also introduced. % to assess the fitness of try-on generations. 

Experiments demonstrate that FitControler can be incorporated into various VTON architectures, providing precise control over garment fit while enhancing the perceptual quality of try-on generations. Ablation studies further show that effective fit control can be achieved with as few as $1{,}000$ training steps % when adapting to other VTON models, 
or with only $1{,}000$ training samples, % when extending to other garment fits, 
highlighting % its
% strong
good training and sample efficiency.

% To our knowledge, our work is the first attempt that systematically investigates fit-aware VTON, from data and methods to metrics and design considerations,
% % which points out a new direction for developing more controllable, realistic, and user-oriented VTON systems.
% which charts a path toward % empowering diverse
% accurate fit control over diverse garments with VTON models.

% \begin{figure}
%   \centering
%   \includegraphics[width=\linewidth]{images/introduction}
%   \caption{\textbf{Comparison between existing VTON models and our proposed FitControler.} (a) Existing models produce only a fixed fit, often leading to unnatural results due to mismatched fit. (b) With FitControler, the same inputs can produce try-on results with customized fits such as \emph{slim}, \emph{regular}, and \emph{loose} T-shirts, which
%   better matches the overall style and user preference.}
%   \label{fig:intro}
% \end{figure}

%------------------------------------------------------------------------
\section{Related Work}
\label{sec:related work}

Our work is related to image-based VTON, customized VTON, and controllable image generation.

\vspace{-10pt}
\subsubsection{Image-based Virtual Try-On.} Image-based VTON aims to transfer a target garment onto a person and synthesize photorealistic try-on images.
Conventional methods~\cite{han2018viton,wang2018toward,yu2019vtnfp,choi2021viton,lee2022high} typically follow a two-stage pipeline by
i) warping the garment to align with the target human pose and  
ii) compositing try-on images using a generative model~\cite{goodfellow2020generative}. 
These approaches, however, can suffer from poor geometric warping under large deformations and unrealistic garment rendering due to the limited generation quality.

Recent diffusion-based methods commonly leverage attention mechanisms to model garment-person interactions, reducing reliance on explicit warping.
They differ mainly in how garment details are captured. For instance, LaDI-VTON~\cite{morelli2023ladi} and DCI-VTON~\cite{gou2023taming} encode garments using CLIP~\cite{radford2021learning}, but CLIP embeddings fail to preserve fine-grained textures due to their focus on high-level semantics. To address this, StableVITON~\cite{kim2024stableviton} employs a ControlNet-like~\cite{zhang2023adding} architecture to capture garment features, enhancing detail preservation. Following TryOnDiffusion~\cite{zhu2023tryondiffusion}, mainstream approaches~\cite{choi2024improving,xu2025ootdiffusion,zhou2025learning,wan2025incorporating,jiang2024fitdit} repurpose the denoising backbone to extract garment features, further improving texture fidelity. Although effective, this dual-branch design significantly increases training and inference costs. TPD~\cite{yang2024texture} and CatVTON~\cite{chong2024catvton} streamline the pipeline by jointly encoding garment and person features within a single U-Net, boosting both efficiency and fidelity.
Despite these advances, they overlook the wearing style, often leading to unnatural results. In particular, the inconsistency of garment fit---such as mismatched tightness between tops and bottoms---can significantly degrade realism. To address this, we explicitly model the garment fit to enable fit-aware VTON.

\vspace{-10pt}
\subsubsection{Customized Virtual Try-On.} 
Due to personalized demands, some work has also explored customized VTON. Landmark-based methods~\cite{yan2023linking,chen2023size,li2024controlling,chen2024wear} enable control over local wearing styles (e.g., sleeve rolling or hem tucking), but typically require manual landmark adjustment. 
COTTON~\cite{chen2023size} removes manual intervention and supports garment length editing, yet overlooks ease allowance, which is critical for proper fit. 
Text-based approaches~\cite{zhu2024m,li2024anyfit,kim2024promptdresser} synthesize try-on images from textual descriptions, offering potential attribute control. PromptDresser~\cite{kim2024promptdresser} demonstrates impressive text-driven editing capabilities; however, its control over garment fit remains implicitly entangled with language semantics and pixel-level inpainting, lacking explicit modeling of body–garment spatial alignment.
Furthermore, the controllability of these approaches is mainly applicable to specific model architectures---even text-based control cannot be 
applied to all diffusion-based models (\eg, CatVTON~\cite{chong2024catvton} drops text input). In this work, we consider garment fit as a holistic, spatially grounded attribute and design a dedicated plug-in for modern diffusion-based VTON models.

%controlnet/T2I-Adapter/IP-Adapter/unicontrol/controlnet++/Unicontrolnet
%
\vspace{-10pt}
\subsubsection{Controllable Image Generation.} 
Recently, notable progress in controllable image generation~\cite{qin2023unicontrol,zhao2023uni,xuctrlora} has enabled conditional guidance such as segmentation maps, canny edges, or depth maps, without retraining text-to-image diffusion models~\cite{nichol2021glide,rombach2022high,peebles2023scalable}. For example, ControlNet~\cite{zhang2023adding} only trains an auxiliary encoder mirroring the structure of the denoising model, and T2I-Adapter~\cite{mou2024t2i} introduces a lightweight network for the same purpose.
These methods, however, focus solely on guiding generation from a conditional image, without considering how  
the condition is acquired. As a result, they typically feature an additional encoder to process the conditional input. In contrast, our approach generates the conditional image itself, allowing us to repurpose features from the generator, which simplifies model design and improves efficiency. 

\begin{figure*}[t]
  \centering
  \includegraphics[width=\linewidth]{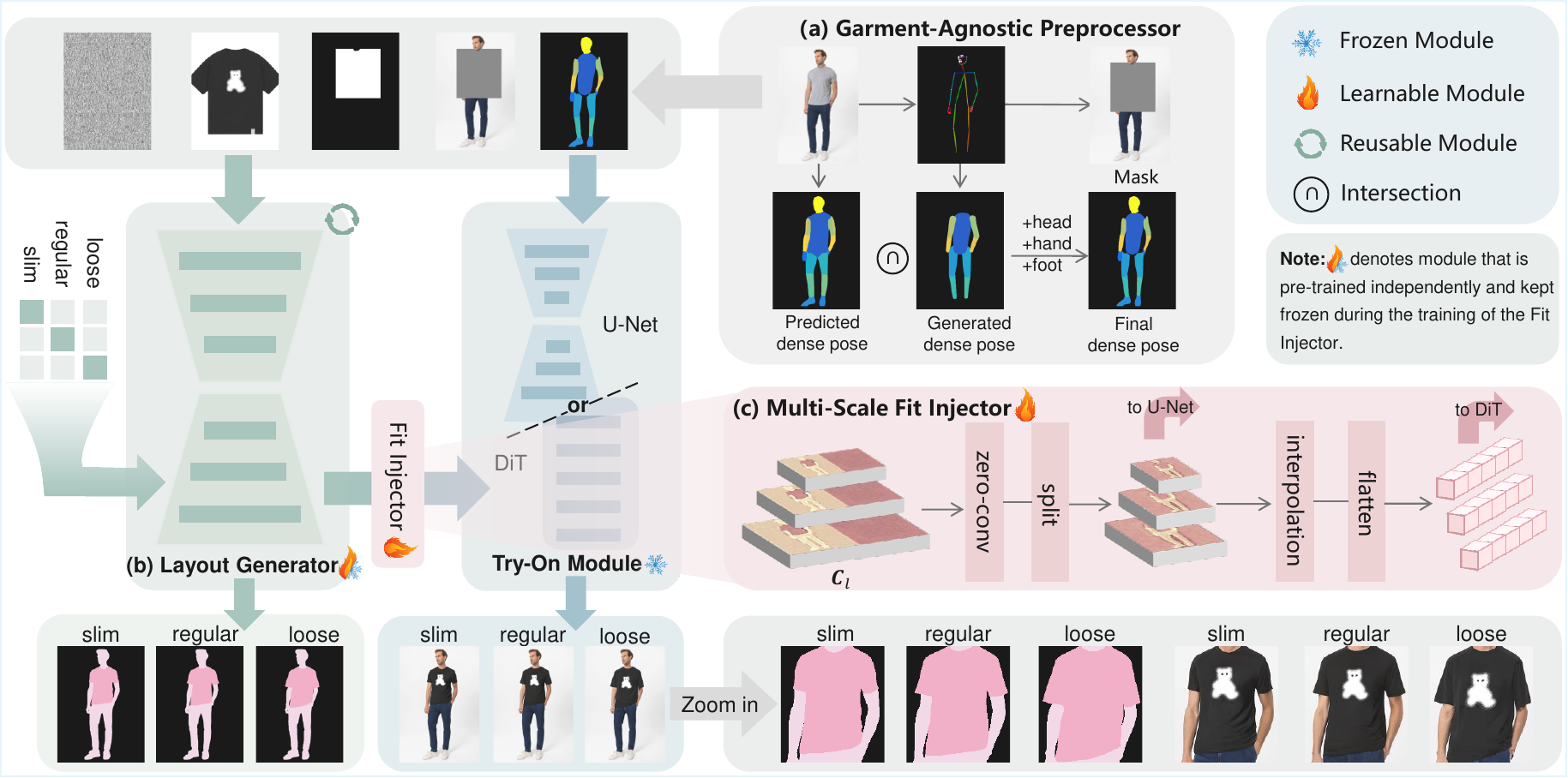}
    \caption{\textbf{Overview of FitControler.} The person image is first processed by (a) the \textit{garment-agnostic preprocessor} to extract the mask and dense pose. These are concatenated with the noise map and garment image---along both channel and spatial dimensions as is done in CatVTON~\cite{chong2024catvton}---before being fed into (b) the \textit{fit-aware layout generator} to produce a fit-sensitive segmentation map. The layout features are then delivered by (c) the \textit{multi-scale fit injector} to the base VTON models via the ControlNet~\cite{zhang2023adding} interface.  The layout generator is pretrained and remains frozen when integrating FitControler into different VTON models, where only the fit injector requires model-specific training.}
  \label{fig:method}
\end{figure*}

%------------------------------------------------------------------------
\section{FitControler}

We begin with the problem setup and then present the key technical details of FitControler.

\subsection{Problem Setup}

Given a person image $\bm{x}_p$ and a garment image $\bm{x}_g$, fit-aware VTON aims to synthesize a try-on image $\bm{x}_{tr}$ according to a % user-specified
target fit prompt $\bm{l}$, where $\bm{x}_p,\bm{x}_g,\bm{x}_{tr} \in \mathbb{R}^{H \times W \times 3}$, with $H$ being the image height and $W$ the image width, and $\bm{l}$ can be of various forms, such as text or categorical labels.

A common paradigm~\cite{chen2024wear,kim2024promptdresser} is to design a dedicated try-on model $\mathcal{T}$ that accepts $\bm{l}$ as an additional input such that
\begin{equation}
\bm{x}_{tr}=\mathcal{T}(\mathcal{P}(\bm{x}_p), \bm{x}_g, \bm{l}) \,,
\end{equation}
where $\mathcal{P}(\bm{x}_p)$ is a preprocessor that generates garment-agnostic representations (\eg, masked images)~\cite{choi2021viton} from $\bm{x}_p$. This paradigm, however, tightly couples the fit control mechanism with a certain 
model and cannot easily integrate into other VTON models.

Unlike the common paradigm, our idea is to design a plug-in $\mathcal{F}$ that is compatible with existing VTON models. This requires addressing two key problems: i) how to encode the fit prompt $\bm{l}$ into the fit feature $\bm{C}_l$ with clear fit awareness and ii) how to enable try-on models to effectively decode $\bm{C}_l$. 
We formulate the two problems as
\begin{align}
\bm{C}_l &= \mathcal{F}(\mathcal{P}(\bm{x}_p), \bm{x}_g, \bm{l}) \,, \\
\bm{x}_{tr} &= \mathcal{T}(\mathcal{P}(\bm{x}_p), \bm{x}_g, \bm{C}_l) \,.
\end{align}

\subsection{FitControler Overview}

Fig.~\ref{fig:method} shows the technical pipeline of FitControler. 
FitControler instructs the fit with categorical labels and features 
i) a \textit{garment-agnostic preprocessor} utilizing human keypoints and body proportions to generate a square-shaped mask and a simulated dense pose;
ii) a \textit{fit-aware layout generator} producing the fit feature that encodes the body-garment spatial relation, conditioned on the fit label; and
iii) a \textit{multi-scale fit injector} mapping the fit feature compatible with try-on models and injects it via the ControlNet~\cite{zhang2023adding} interface.

\subsection{Garment-Agnostic Preprocessor}
\label{sec:mask}
\begin{figure*}[t]
  \centering
  \includegraphics[width=\linewidth]{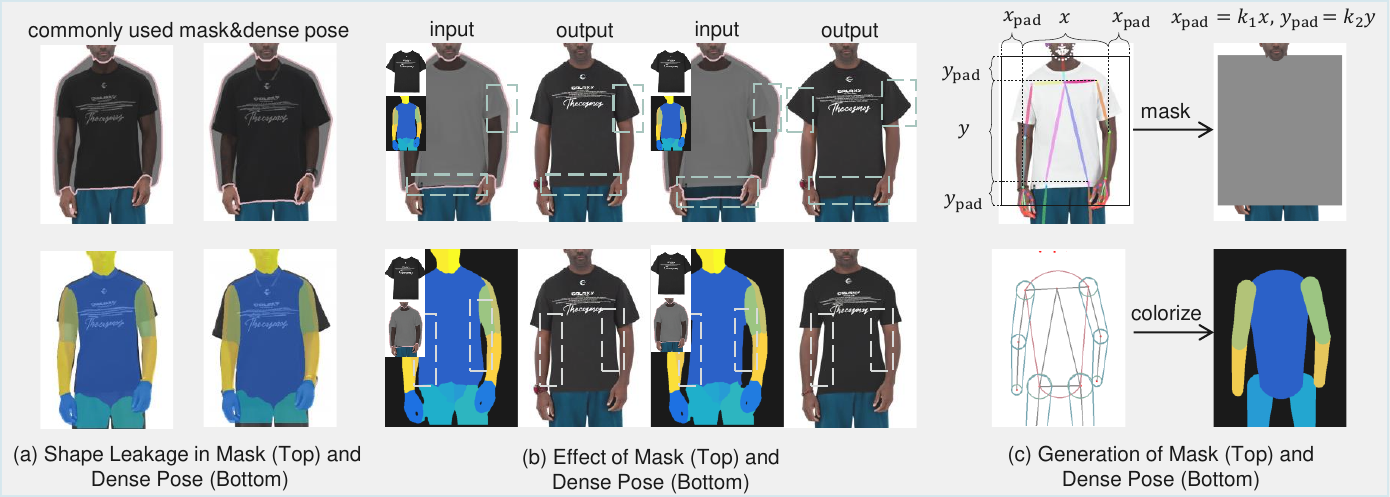}
  \caption{\textbf{Impact of shape leakage in mask and dense pose.} 
  (a) Commonly used masks and dense poses implicitly encode the original garment boundaries, 
  which (b) biases the model to follow these cues when rendering garment. To address this, (c) our preprocessor reconstructs them from human keypoints to form standardized representations. See supplementary material for the ablation on this preprocessor.}
  \label{fig:mask_pose}
\end{figure*}

Garment-agnostic representations are standard inputs to VTON models that aim to reduce the cost of data collection. They enable training with ($\tt garment$, $\tt try\mbox{-}on\ image$) pairs under reconstruction-based supervision. %, instead of using ($\tt person$, $\tt garment$, $\tt try\mbox{-}on\ image$) triplets.
However, commonly used representations such as the mask and the dense pose (Fig.~\ref{fig:mask_pose}(a)) often preserve the original garment contour, because the mask is typically generated along the garment boundary, and DensePose~\cite{guler2018densepose} tends to misinterpret garment edges as body boundaries. 
Consequently, the model is biased to render garments following these leaked cues (Fig.~\ref{fig:mask_pose}(b)), indicating that the standard representations do not meet the need of fit-aware synthesis. To address this, we redesign the garment-agnostic preprocessor $\mathcal{P}(\bm{x}_p)$ to eliminate garment shape leakage.

\vspace{-10pt}
\subsubsection{Mask.} Unlike prior work~\cite{choi2021viton} that estimates masks from detected garment regions, we derive masks from human keypoints to avoid garment contour leakage. Specifically, for the upper body, we define $x$ as the horizontal span between the minimum and maximum coordinates of the shoulder, elbow, and wrist joints, and $y$ as the vertical span between the shoulder and hip keypoints. The mask is generated as in Fig.~\ref{fig:mask_pose}(c), where two empirical padding ratios $k_1=0.6$ and $k_2=0.25$ to fully cover the clothing region while minimizing unnecessary background.
For the lower body, we construct the mask based on the hip, knee, and ankle keypoints, with $k_1=0.5$ and $k_2=0.2$. 

\vspace{-10pt}
\subsubsection{DensePose.} To mitigate inaccuracies in the predicted dense pose, we synthesize a standard-stature dense pose based on human keypoints and canonical body proportions. 
According to Fig.~\ref{fig:mask_pose}(c), the torso is approximated by a quadrilateral defined by shoulder and hip joints, with convex arcs at the top and bottom to yield natural contours. Limbs are constructed by connecting circles placed at major joints (shoulder--elbow--wrist and hip--knee--ankle), whose diameters are proportional to body height (\eg, $0.06$, $0.048$, $0.033$ for the arm; $0.09$, $0.055$, $0.03$ for the leg). Body height is estimated to be of $3.2\times$ torso height or $2.3\times$ leg length. Finally, the synthesized map is intersected with the predicted dense pose to constrain the spatial layout.

Notably, what matters here is the general design principle of constructing garment-agnostic masks and dense poses to eliminate shape leakage; the specific empirical parameters are not critical to the overall effectiveness.

\subsection{Fit-Aware Layout Generator}
\label{sec:Segmentation}
The fit-aware layout generator produces a body-garment layout conditioned on the fit label $\bm{l}$. We represent the layout as a segmentation map and repurpose the U-Net from Stable Diffusion~\cite{rombach2022high} to achieve this task. 
The rich semantic priors from the pretrained U-Net can benefit layout prediction.    

Formally, given a person image $\bm{x}_p$ and a garment image $\bm{x}_g$, we first derive the mask $\bm{m}$, the masked person image $\bm{x}_m$, and the dense pose $\bm{x}_d$ via $\mathcal{P}(\bm{x}_p)$. 
Since the U-Net operates in the latent space of a VAE~\cite{kingma2013auto}, these inputs are first encoded by the VAE encoder $\mathcal{E}(\cdot)$, with the mask $\bm{m}$ downsampled to the corresponding resolution. A Gaussian noise map $\bm{z}_T$ is further sampled as the stochastic input.
The input to the generator is a $13$-channel tensor
\begin{equation}
\mathcal{X} = \bm{z}_T \mathbin{||} [\bm{m}\oplus \bm{0}] \mathbin{||} [\mathcal{E}(\bm{x}_m)\oplus \mathcal{E}(\bm{x}_g)] \mathbin{||} [\mathcal{E}(\bm{x}_d)\oplus \bm{0})] \,,
\end{equation}
along with the one-hot encoded fit label $\bm{l}$, where $\bm{0}$ is a zero tensor used for spatial alignment, $\oplus$ denotes spatial concatenation, and $\mathbin{||}$ channel-wise concatenation, following CatVTON~\cite{chong2024catvton}. Then the layout generator $\mathcal{G_S}$ produces
\begin{equation}
\bm{S}_l, \bm{C}_l = \mathcal{G_S}(\mathcal{X}, \bm{l}) \,,
\end{equation}
where $\bm{S}_l$ is a three-class segmentation map ($\tt background$, $\tt body$, and $\tt garment$), and $\bm{C}_l$ is the fit-aware features aggregated from multi-scale intermediate decoder features before ResNet blocks.
To condition $\bm{l}$, we follow the prior controlled image generation pipeline~\cite{preechakul2022diffusion,peebles2023scalable} and inject it through feature-level modulation using FiLM~\cite{perez2018film} in each ResNet block of the decoder, which amounts to
\begin{equation}
\bm{h}_s \leftarrow \gamma(\bm{l}) \cdot \frac{\bm{h}_s-\mu}{\sigma} + \beta(\bm{l}) \,,
\end{equation}
where $\bm{h}_s$ denotes intermediate features, $\mu$ and $\sigma$ are feature-wise statistics, and $\gamma(\cdot)$ and $\beta(\cdot)$ are learnable projections. Notably, we freeze the U-Net encoder to preserve its pretrained semantic knowledge. Additional implementation details are provided in the supplementary material.

\subsection{Multi-Scale Fit Injector}
\label{sec:adapter}
The fit injector transforms the multi-scale layout features $\bm{C}_l$ into a compatible representation that is acceptable by VTON models.

At each scale, a zero-initialized convolution layer is first applied to align the layout features with try-on features.
To ensure spatial consistency, we apply operations such as splitting and interpolation to match the resolution of the try-on model. For U-Net based models (\eg, Leffa~\cite{zhou2025learning}), resolution alignment can be achieved by simple splitting operations. In contrast, DiT-based models (\eg, FitDiT~\cite{jiang2024fitdit}) require interpolated feature maps of a uniform resolution, followed by flattening. 

During the training of the injector, both the layout generation module and the try-on module remain frozen, and only the injector parameters are optimized using the loss
\begin{equation}
\mathcal{L}=\mathbb{E}_{\zeta,t,\epsilon,C_l}\left[ \left \| \epsilon-\epsilon_\theta (\bm{\zeta},t,\tau_{\phi} (\bm{C}_l))\right \|^{2}_{2} \right] \,,
\label{eq:loss}
\end{equation}
where $\tau_\phi$ is the injector, $t$ is the diffusion timestep, $\epsilon$ is the noise, and $\bm{\zeta}$ denotes the try-on input. For example, in Leffa, $\bm{\zeta}=[\mathcal{E}(\bm{x}_g),\, \bm{z}_t \mathbin{||} \mathcal{E}(\bm{x}_m) \mathbin{||} \mathcal{E}(\bm{x}_d) \mathbin{||} \bm{m}]$.

\section{Fit Consistency Metric}

Evaluating how well a try-on image adheres to a specific fit is non-trivial, because
fit differences manifest in not only realism but also subtle garment contours. This requires explicit contour-aware metrics to compare fit consistency. 
Concretely, for each source image, we generate a try-on image conditioned on the same fit label, extract the garment contours from both, and measure their similarity. We adopt two complementary shape metrics: Hu moments (Hu)~\cite{hu1962visual} and Hausdorff distance (Hd)~\cite{huttenlocher2002comparing}.

\vspace{-10pt}
\subsubsection{Hu Moments.} Hu delineates the shape globally. Given a binary contour mask $I(x,y)$, the central moment of order $(p,q)$ is first computed by
\begin{equation}
\mu_{pq} = \sum_x \sum_y (x-\bar{x})^p (y-\bar{y})^q I(x,y) \,,
\end{equation}
where $(\bar{x},\bar{y})$ is the centroid. % The moments are
It is then normalized to
\begin{equation}
\eta_{pq} = \frac{\mu_{pq}}{\mu_{00}^{r}}, \quad r = \frac{p+q}{2}+1 \,.
\end{equation}
Hu takes the form $\Phi(I) = [\phi_1, \dots, \phi_7]$, where each $\phi_i$ is a specific combination of $\eta_{pq}$'s, \eg, $\phi_1=\eta_{20}+\eta_{02}$. 
Hu captures the spatial distribution of contour pixels via moment statistics, providing a compact yet informative representation of global shape. In practice, we compute the Hu distance between two images as a single scalar.
The closer the distance between \(\Phi(I_{\text{gen}})\) and \(\Phi(I_{\text{src}})\) is, the better the garment contour 
matches, where \(I_{\text{gen}}\) and \(I_{\text{src}}\) refer to the generated and source contours, respectively.

\begin{figure}[t]
  \centering
  \includegraphics[width=\linewidth]{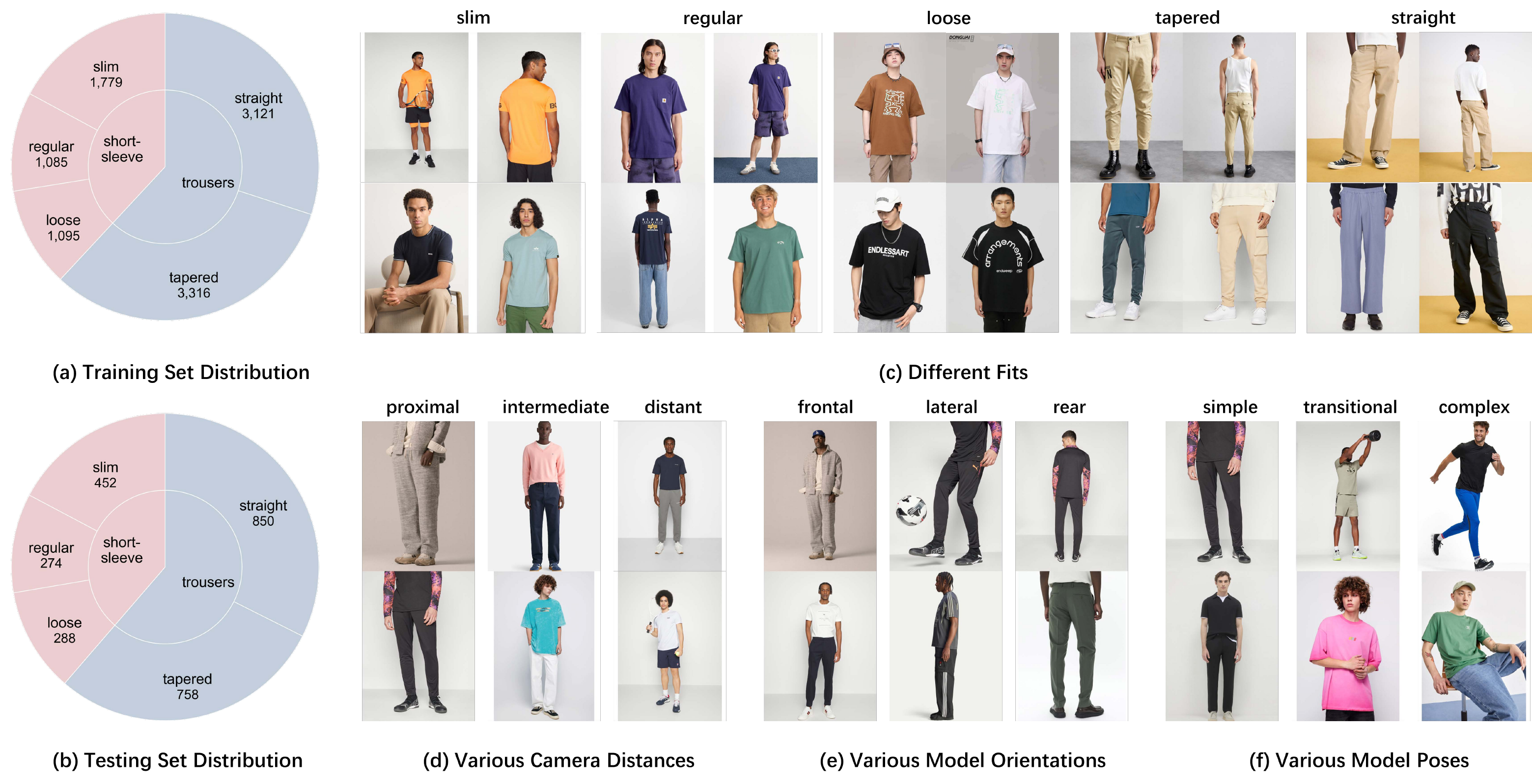}
  \caption{\textbf{Overview of Fit4Men.} (a) and (b) present the statistical distributions of the training and testing sets. (c) Examples of five garment fit types supporting fit-aware learning. (d) shows samples captured at different camera distances, while (e) and (f) depict variations in model orientations and poses, respectively, ensuring the diversity of Fit4Men.}
  \label{fig:dataset_overview}
\end{figure}

\vspace{-10pt}
\subsubsection{Hausdorff Distance.} Hd captures local geometric deviations. Given two contour point sets $A$ and $B$, the Hausdorff distance $d_H(A,B)$ between $A$ and $B$ takes the form
\begin{equation}
d_H(A,B) = \max \left \{ \sup_{a \in A} \inf_{b \in B} d(a,b), \ \sup_{b \in B} \inf_{a \in A} d(b,a) \right \} \,,
\end{equation}
where $d(a,b)$ is the Euclidean distance between points $a$ and $b$ in our case.
Hd measures the maximum nearest-neighbor distance between contour points, highlighting mismatches such as the misalignment of sleeve or hem.

With Hu and Hd, our fit consistency metrics take both global shape coherence and local fidelity into account. Lower values of both metrics indicate better performances.

\section{Fit4Men Dataset}
\label{sec:Fit-Annotated Dataset}

We construct \textbf{Fit4Men}, a medium-resolution ($768 \times 1024$) dataset specifically designed for fit-aware VTON. 
As illustrated in Fig.~\ref{fig:dataset_overview}, it comprises $5{,}000$ pairs of men's short-sleeve shirts categorized into three fit types ($\tt slim$, $\tt regular$, and $\tt loose$) along with $8{,}000$ pairs of men's trousers featuring two fit types ($\tt tapered$ and $\tt straight$) sourced from e-commerce platforms.\footnote{\href{https://zalando.com/}{Zalando}, \href{https://www.taobao.com}{Taobao}, and \href{https://global.musinsa.com/}{Musinsa}.} 
The dataset is split into training and testing sets at a ratio of $4{:}1$. 
Regarding annotations, human keypoints are extracted using DWPose~\cite{yang2023effective}, while DensePose~\cite{guler2018densepose} and Sapiens~\cite{khirodkar2024sapiens} are employed to generate dense poses and ground-truth segmentation maps, respectively.
To replicate realistic try-on scenarios, Fit4Men incorporates variations in camera distance, model orientation, and pose complexity.

\vspace{-10pt}
\subsubsection{Camera Distance.}
Fit4Men encompasses a broad range of camera distances frequently encountered in fashion photography.
These distances are classified into three levels consisting of proximal, intermediate, and distant views.
Proximal views capture specific body regions (\eg, upper or lower body), whereas intermediate views depict the full body within the frame, and distant views portray the subject with significant spatial separation from the camera. 
This variation enhances robustness against framing discrepancies.

\vspace{-10pt}
\subsubsection{Model Orientation.}
The dataset features subjects captured from diverse orientations, specifically frontal, lateral, and rear views. 
Such viewpoint diversity mirrors real-world catalog photography while enhancing generalization to arbitrary viewing angles during the virtual try-on process.

\vspace{-10pt}
\subsubsection{Pose Complexity.}
Fit4Men addresses three levels of pose complexity including simple, moderate, and complex categories.
Simple poses represent upright standing positions characterized by an absence of limb occlusion.
Moderate poses involve partial self-occlusion (\eg, crossed arms or bent legs).
Complex poses encompass dynamic or non-standing states including sitting or running.
This design facilitates the evaluation of fit consistency under diverse body configurations.

\section{Results and Discussion}

Here we present our results, and discussion.

\subsection{Experimental Setup}
\label{sec:experiment}

% \subsubsection{Dataset.} Due to the absence of fit-aware VTON dataset, we introduce Fit4Men, a medium-resolution ($768 \times 1024$) dataset featuring $5,000$ pairs of men's short-sleeve T-shirts and $8,000$ pairs of men's trousers collected from e-commerce platforms.\footnote{Zalando, Taobao, and Musinsa}
% % \footnote{Zalando (\href{https://www.zalando.de}{https://www.zalando.de}), Taobao (\href{https://www.taobao.com}{https://www.taobao.com}), and Musinsa (\href{https://www.musinsa.com}{https://www.musinsa.com})}
% T-shirts are labeled with three fit types ($\tt slim$, $\tt regular$, and $\tt loose$), and trousers with two ($\tt tapered$ and $\tt straight$). 
% The number of sample pairs is balanced across different fits. The dataset encompasses diverse camera distances and model
% poses. 
% Further details are provided in the supplementary material.

% \vspace{-10pt}

\subsubsection{Implementation Details.}
We first train the layout generator using the cross-entropy loss. Subsequently, for each try-on model, the fit injector is trained for $7{,}500$ steps following the configuration described in Sec.~\ref{sec:adapter}. All models are optimized using AdamW~\cite{loshchilov2017decoupled} with a batch size of $64$ and a learning rate of $1 \times 10^{-5}$ on Fit4Men. Training and evaluation are performed at $384 \times 512$ resolution with FP16 precision on four NVIDIA A6000 GPUs. Additional training details are provided in the supplementary material.

\vspace{-10pt}
\subsubsection{Experimental Protocol.} To demonstrate generality, we integrate FitControler into five state-of-the-art VTON models: StableVITON~\cite{kim2024stableviton}, IDM-VTON~\cite{choi2024improving}, CatVTON (and its FLUX variant)~\cite{chong2024catvton}, Leffa~\cite{zhou2025learning}, and FitDiT~\cite{jiang2024fitdit}, covering diffusion backbones from SD1.5~\cite{rombach2022high} and SDXL~\cite{podell2023sdxl} to SD3~\cite{esser2024scaling} and FLUX.1~\cite{flux2024}. Each model is evaluated on Fit4Men with and without FitControler, under both paired and unpaired settings. 
The baseline VTON models use their developed preprocessing pipelines for generating garment-agnostic representations. Following prior work~\cite{chong2024catvton,zhang2025boow}, we also report FID~\cite{heusel2017gans}, KID~\cite{binkowski2018demystifying}, SSIM~\cite{wang2004image}, and LPIPS~\cite{zhang2018unreasonable} to assess image quality in the paired setting, and FID/KID in the unpaired setting. Both settings also report Hu and Hd to evaluate fit consistency. 

\begin{table}[!t]
\centering
\tiny
\caption{Analysis of fit consistency metrics. 
\textbf{pd} is the generated image, and \textbf{gt} the source image. 
Best performance is in \textbf{boldface}.}
\label{tab:metrics}

\vspace{-12pt}

\begin{subtable}[t]{0.58\linewidth}
\centering

\caption{Upper-body garments.}
\label{tab:metrics_upper}
\addtolength{\tabcolsep}{2pt}
\renewcommand{\arraystretch}{1.1}
\vspace{-8pt}
\resizebox{\textwidth}{!}{
\begin{tabular}{lcccccc}
\toprule
\multirow{2}*{\diagbox{pd}{gt}} 
& \multicolumn{2}{c}{slim}
& \multicolumn{2}{c}{regular}
& \multicolumn{2}{c}{loose} \\
\cmidrule(r){2-3}\cmidrule(r){4-5}\cmidrule(r){6-7}
& Hu & Hd & Hu & Hd & Hu & Hd \\
\midrule
slim    & \textbf{0.32} & \textbf{6.46} & 0.43 & 7.60 & 0.67 & 14.34 \\
regular & 0.41 & 7.65 & \textbf{0.39} & \textbf{6.35} & 0.54 & 11.16 \\
loose   & 0.58 & 11.61 & 0.47 & 9.24 & \textbf{0.43} & \textbf{7.34} \\
\bottomrule
\end{tabular}
}
\end{subtable}
\hfill
\begin{subtable}[t]{0.4\linewidth}
\centering
\caption{Lower-body garments.}
\label{tab:metrics_lower}
\addtolength{\tabcolsep}{1pt}
\renewcommand{\arraystretch}{1.375}
\vspace{-8pt}
\resizebox{\textwidth}{!}{
\begin{tabular}{lcccc}
\toprule
\multirow{2}*{\diagbox{pd}{gt}} 
& \multicolumn{2}{c}{tapered}
& \multicolumn{2}{c}{straight} \\
\cmidrule(r){2-3}\cmidrule(r){4-5}
& Hu & Hd & Hu & Hd \\
\midrule
tapered  & \textbf{0.62} & \textbf{6.43} & 1.74 & 13.14 \\
straight & 1.47 & 12.83 & \textbf{0.91} & \textbf{9.47} \\
\bottomrule
\end{tabular}
}
\end{subtable}

\vspace{-6pt}
\end{table}

\begin{figure}[!t]
  \centering
  \includegraphics[width=\linewidth]{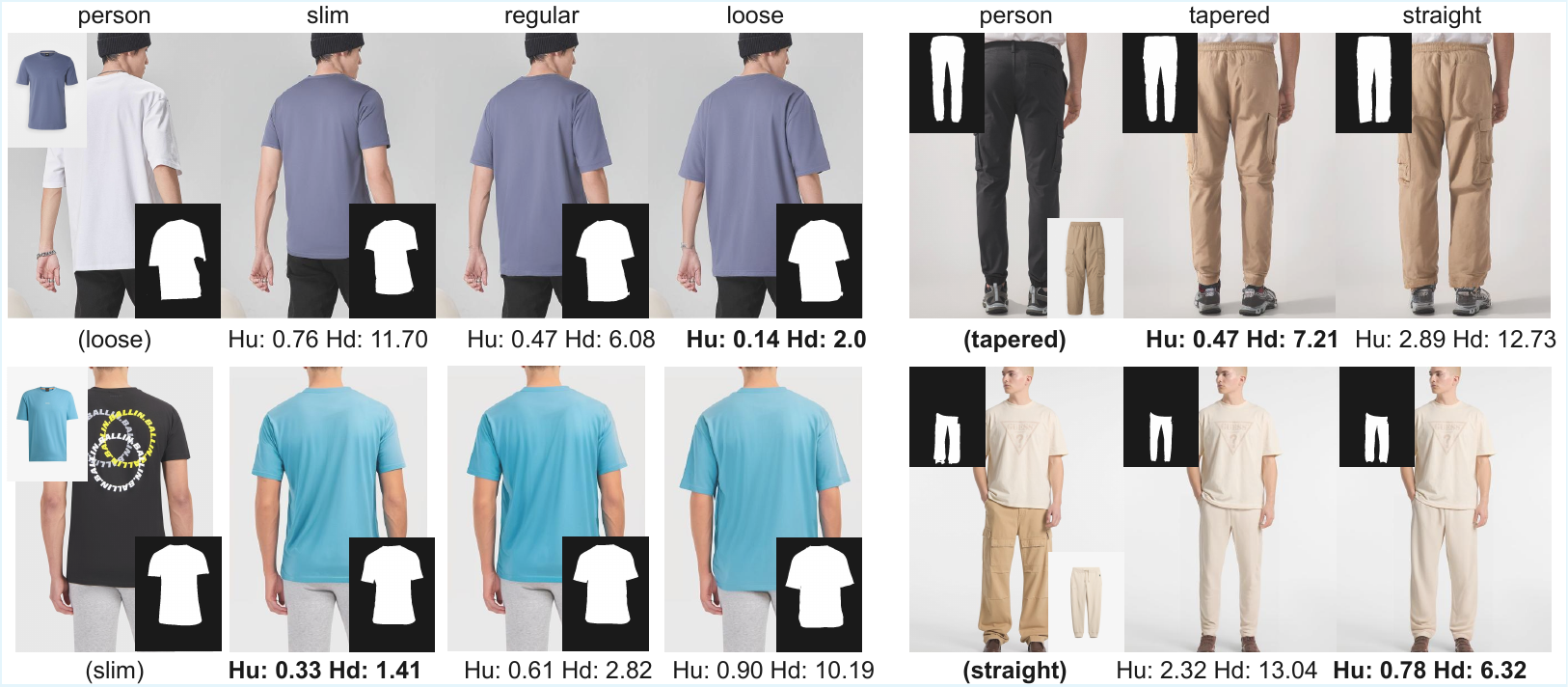}
  \vspace{-15pt}
  \caption{\textbf{Correlation analysis of fit consistency metrics.} For each target fit, three types of fits (slim, regular, and loose) are generated and compared.}
  \label{fig:metrics}
  \vspace{-10pt}
\end{figure}

\subsection{Sanity Check on Fit Consistency Metrics}
Before reporting Hu and Hd, we first perform a sanity check to confirm their soundness in assessing fit consistency.  
We analyze both qualitative and quantitative results on Fit4Men.
For the qualitative analysis, we generate three different fits from a person-garment pair and compute the Hu and Hd values relative to the original image for short-sleeve shirts.
For the quantitative analysis, we partition the the short-sleeve test set into three subsets according to their fit labels and generate VTON results w.r.t. the three different fits for each subset.
An analogous protocol is applied to trousers.
According to Table~\ref{tab:metrics} and Fig.~\ref{fig:metrics}, both metrics achieve the best values when the generated fit matches the original one, and increase progressively with larger fit differences (\eg, loose vs. slim). This confirms Hu and Hd 
are suitable for evaluating fit consistency.

\begin{table}[t]
\centering
\footnotesize
\renewcommand{\arraystretch}{1.1}
\addtolength{\tabcolsep}{1pt}
\newcommand{\imp}[1]{\raisebox{-0ex}{\hspace{1pt}\scriptsize\textcolor{ForestGreen}{#1}}} % improvement (下降指标)
\newcommand{\impup}[1]{\raisebox{-0ex}{\hspace{1pt}\scriptsize\textcolor{NiceRed}{#1}}} % 上升指标（红色）
\caption{Performance of VTON models with/without FitControler on Fit4Men. Relative improvements are colored in \textcolor{NiceRed}{red} $\uparrow$ and \textcolor{ForestGreen}{green} $\downarrow$.}
\vspace{-5pt}
% \begin{tabular}{lcccccccccc}
\resizebox{\textwidth}{!}{
\begin{tabular}{lllllllllll}
\toprule
\multirow{2}*{Method} & \multicolumn{6}{c}{Paired} & \multicolumn{4}{c}{Unpaired}\\
\cmidrule(r){2-7}\cmidrule(r){8-11}
&FID$\downarrow$ & KID$\downarrow$ & SSIM$\uparrow$ & LPIPS$\downarrow$ & Hu$\downarrow$ & Hd$\downarrow$
&FID$\downarrow$ & KID$\downarrow$ & Hu$\downarrow$ & Hd$\downarrow$ \\
\midrule
\textbf{Short-sleeve} \\
StableVITON~\cite{kim2024stableviton}
&16.93 &4.34 &0.853 &0.0653 &0.57 &9.90
&21.97 &5.75 &0.66 &10.35 \\
\rowcolor{gray!10}+FitControler
&\textbf{14.39}\imp{(15.0\%)} 
&\textbf{2.53}\imp{(41.7\%)} 
&\textbf{0.874}\impup{(2.5\%)} 
&\textbf{0.0634}\imp{(2.9\%)} 
&\textbf{0.42}\imp{(26.3\%)} 
&\textbf{7.63}\imp{(22.9\%)} 
&\textbf{19.96}\imp{(9.1\%)} 
&\textbf{4.81}\imp{(16.3\%)} 
&\textbf{0.50}\imp{(24.2\%)} 
&\textbf{8.54}\imp{(17.5\%)} \\
IDMVTON~\cite{choi2024improving}
&15.99 &2.90 &0.852 &0.0647 &0.55 &9.03
&20.83 &3.37 &0.67 &10.39 \\
\rowcolor{gray!10}+FitControler
&\textbf{14.69}\imp{(8.1\%)} 
&\textbf{1.92}\imp{(33.8\%)} 
&\textbf{0.866}\impup{(1.6\%)} 
&\textbf{0.0592}\imp{(8.5\%)} 
&\textbf{0.45}\imp{(18.2\%)} 
&\textbf{7.54}\imp{(16.5\%)} 
&\textbf{19.11}\imp{(8.3\%)} 
&\textbf{2.19}\imp{(35.0\%)} 
&\textbf{0.51}\imp{(23.9\%)} 
&\textbf{8.81}\imp{(15.2\%)} \\
CatVTON~\cite{chong2024catvton}
&14.28 &2.02 &0.864 &0.0574 &0.57 &9.17
&19.09 &3.35 &0.64 &10.58 \\
\rowcolor{gray!10}+FitControler
&\textbf{13.01}\imp{(8.9\%)} 
&\textbf{1.06}\imp{(47.5\%)} 
&\textbf{0.873}\impup{(1.0\%)} 
&\textbf{0.0552}\imp{(3.8\%)} 
&\textbf{0.44}\imp{(22.8\%)} 
&\textbf{7.13}\imp{(22.2\%)} 
&\textbf{18.29}\imp{(4.2\%)} 
&\textbf{2.07}\imp{(38.2\%)} 
&\textbf{0.49}\imp{(23.4\%)} 
&\textbf{8.64}\imp{(18.3\%)} \\
CatVTON-FLUX~\cite{chong2024catvton}
&15.76 &2.54 &0.855 &0.0764 &0.56 &9.41
&20.40 &3.04 &0.65 &10.56 \\
\rowcolor{gray!10}+FitControler
&\textbf{14.44}\imp{(8.4\%)} 
&\textbf{1.39}\imp{(45.3\%)} 
&\textbf{0.860}\impup{(0.6\%)} 
&\textbf{0.0652}\imp{(14.7\%)} 
&\textbf{0.44}\imp{(21.4\%)} 
&\textbf{7.92}\imp{(15.8\%)} 
&\textbf{18.40}\imp{(9.8\%)} 
&\textbf{1.93}\imp{(36.5\%)} 
&\textbf{0.47}\imp{(27.7\%)} 
&\textbf{8.47}\imp{(19.8\%)} \\
FitDiT~\cite{jiang2024fitdit}
&17.40 &4.17 &0.864 &0.0723 &0.65 &12.36
&23.94 &6.82 &0.80 &16.89 \\
\rowcolor{gray!10}+FitControler
&\textbf{12.51}\imp{(28.1\%)} 
&\textbf{0.90}\imp{(78.4\%)} 
&\textbf{0.875}\impup{(1.3\%)} 
&\textbf{0.0510}\imp{(29.5\%)} 
&\textbf{0.38}\imp{(41.5\%)} 
&\textbf{6.77}\imp{(45.2\%)} 
&\textbf{18.48}\imp{(22.8\%)} 
&\textbf{1.86}\imp{(72.7\%)} 
&\textbf{0.46}\imp{(42.5\%)} 
&\textbf{7.94}\imp{(53.0\%)} \\
Leffa~\cite{zhou2025learning}
&14.59 &1.03 &0.861 &0.0668 &0.56 &9.18
&18.77 &2.59 &0.65 &10.47 \\
\rowcolor{gray!10}+FitControler
&\textbf{12.45}\imp{(14.7\%)} 
&\textbf{0.62}\imp{(39.8\%)} 
&\textbf{0.869}\impup{(0.9\%)} 
&\textbf{0.0530}\imp{(20.7\%)} 
&\textbf{0.40}\imp{(28.6\%)} 
&\textbf{7.05}\imp{(23.2\%)} 
&\textbf{17.73}\imp{(5.5\%)} 
&\textbf{1.54}\imp{(40.5\%)} 
&\textbf{0.45}\imp{(30.8\%)} 
&\textbf{7.65}\imp{(26.9\%)} \\

\midrule
\textbf{Trousers} \\
IDMVTON~\cite{choi2024improving}
&15.11 &6.67 &0.854 &0.0677 &1.03 &10.53
&16.12 &5.59 &1.58 &12.58 \\
\rowcolor{gray!10}+FitControler
&\textbf{12.14}\imp{(19.7\%)} 
&\textbf{3.68}\imp{(44.8\%)} 
&\textbf{0.862}\impup{(0.9\%)} 
&\textbf{0.0592}\imp{(12.6\%)} 
&\textbf{0.81}\imp{(21.4\%)} 
&\textbf{9.32}\imp{(11.5\%)} 
&\textbf{14.43}\imp{(10.5\%)} 
&\textbf{3.95}\imp{(29.3\%)} 
&\textbf{1.13}\imp{(28.5\%)} 
&\textbf{10.29}\imp{(18.2\%)} \\
CatVTON~\cite{chong2024catvton}
&11.46 &3.43 &0.852 &0.0637 &0.92 &9.50
&14.19 &3.94 &1.48 &11.92 \\
\rowcolor{gray!10}+FitControler
&\textbf{10.02}\imp{(12.6\%)} 
&\textbf{1.82}\imp{(46.9\%)} 
&\textbf{0.867}\impup{(1.8\%)} 
&\textbf{0.0582}\imp{(8.6\%)} 
&\textbf{0.80}\imp{(13.0\%)} 
&\textbf{8.71}\imp{(8.3\%)} 
&\textbf{12.22}\imp{(13.0\%)} 
&\textbf{1.96}\imp{(50.2\%)} 
&\textbf{1.03}\imp{(30.4\%)} 
&\textbf{9.39}\imp{(21.2\%)} \\
CatVTON-FLUX~\cite{chong2024catvton}
&13.81 &5.61 &0.844 &0.0811 &1.26 &13.38
&16.87 &7.34 &1.67 &15.44 \\
\rowcolor{gray!10}+FitControler
&\textbf{12.15}\imp{(12.0\%)} 
&\textbf{4.13}\imp{(26.4\%)} 
&\textbf{0.851}\impup{(0.8\%)} 
&\textbf{0.0796}\imp{(1.9\%)} 
&\textbf{0.91}\imp{(27.8\%)} 
&\textbf{9.49}\imp{(29.0\%)} 
&\textbf{15.42}\imp{(8.6\%)} 
&\textbf{6.42}\imp{(12.5\%)} 
&\textbf{1.19}\imp{(28.7\%)} 
&\textbf{10.78}\imp{(30.2\%)} \\
FitDiT~\cite{jiang2024fitdit}
&14.00 &4.32 &0.841 &0.0832 &1.26 &13.36
&15.74 &4.30 &1.86 &17.49 \\
\rowcolor{gray!10}+FitControler
&\textbf{9.94}\imp{(28.9\%)} 
&\textbf{1.91}\imp{(55.7\%)} 
&\textbf{0.868}\impup{(3.1\%)} 
&\textbf{0.0593}\imp{(28.7\%)} 
&\textbf{0.75}\imp{(40.5\%)} 
&\textbf{7.59}\imp{(43.1\%)} 
&\textbf{12.67}\imp{(19.5\%)} 
&\textbf{2.39}\imp{(44.4\%)} 
&\textbf{1.00}\imp{(46.2\%)} 
&\textbf{8.90}\imp{(49.1\%)} \\
Leffa~\cite{zhou2025learning}
&10.56 &3.27 &0.850 &0.0720 &1.17 &10.39
&12.55 &4.03 &1.31 &10.63 \\
\rowcolor{gray!10}+FitControler
&\textbf{9.83}\imp{(6.9\%)} 
&\textbf{1.90}\imp{(41.9\%)} 
&\textbf{0.864}\impup{(1.6\%)} 
&\textbf{0.0610}\imp{(15.3\%)} 
&\textbf{0.74}\imp{(36.8\%)} 
&\textbf{7.87}\imp{(24.3\%)} 
&\textbf{11.98}\imp{(4.5\%)} 
&\textbf{2.12}\imp{(47.4\%)} 
&\textbf{0.98}\imp{(25.2\%)} 
&\textbf{8.80}\imp{(17.2\%)} \\

\bottomrule
\end{tabular}
}
\label{tab:compare}
\end{table}

\begin{figure}[!t]
  \centering
  \includegraphics[width=\linewidth]{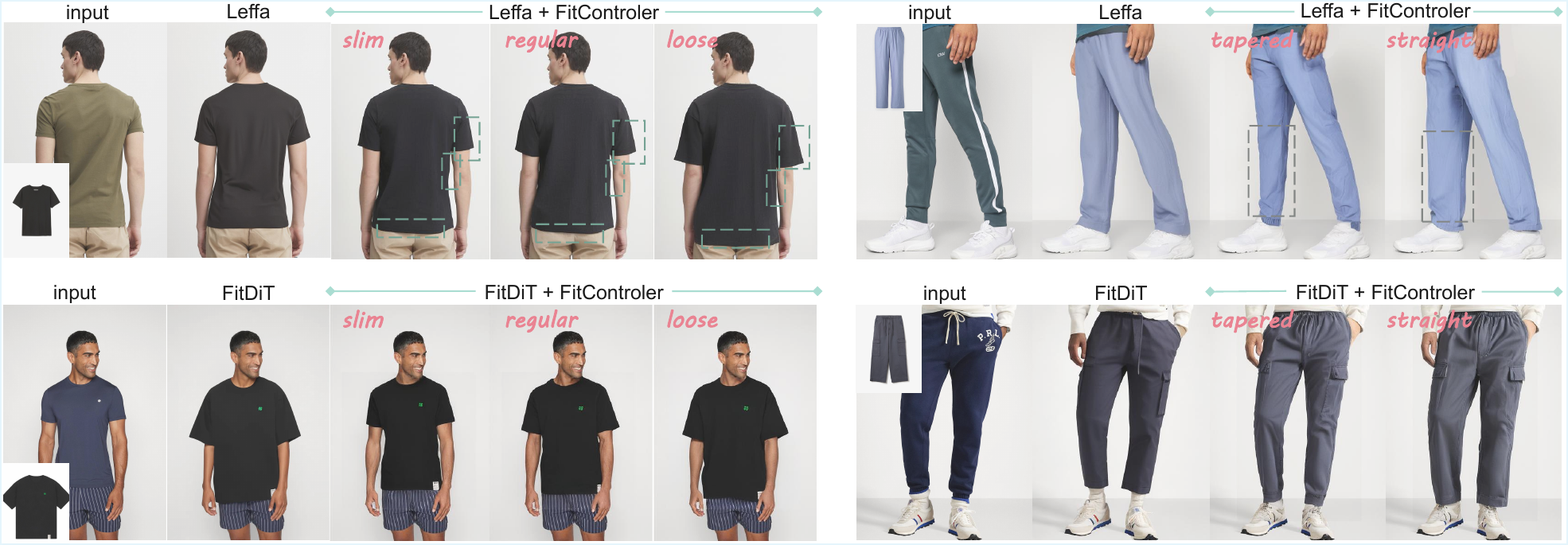}
  \vspace{-15pt}
  \caption{\textbf{Qualitative results of VTON models with FitControler.}
  Regions showing the most prominent fit variations are highlighted with dashed boxes. Additional examples on other VTON models are provided in the supplementary material.}
  \label{fig:qualitative}
\end{figure}

\subsection{Main Results}
Here we compare the performance of state-of-the-art VTON models with and without FitControler on the Fit4Men dataset. Since StableVITON~\cite{kim2024stableviton} provides pretrained weights only for upper-body data, it is only evaluated on the short-sleeve subset.
Quantitative results are shown in Table~\ref{tab:compare}. One can observe that FitControler consistently improves the fit consistency metrics in both Hu and Hd. 
Interestingly, improved garment fits also lead to consistently improved performance over other image quality metrics (FID, KID, SSIM, and LPIPS), which suggests that fit is an essential factor in achieving realism, but has been underexplored in previous works.
Notably, as discussed in Section~\ref{sec:mask}, models such as Leffa~\cite{zhou2025learning} rely on masks and dense poses that partially leak the original garment, leading to limited gains from FitControler. In contrast, FitDiT~\cite{jiang2024fitdit} uses square masks and human keypoints to remove fit cues and thus benefits substantially from FitControler, improving all metrics by large margins.
In addition, we evaluate performance on a regular-fit-only subset of the short-sleeve data. 
The results, provided in the supplementary material, follow trends consistent with Table~\ref{tab:compare}. 
These results confirms that FitControler enables effective fit-aware VTON while enhancing the overall realism via improved fit consistency.

Fig.~\ref{fig:abs} presents the qualitative results of FitControler on CatVTON~\cite{chong2024catvton}, and Fig.~\ref{fig:qualitative} further shows the results on Leffa and FitDiT.
For short-sleeve T-shirts, the generated fits exhibit clear differences in sleeve tightness and torso contour while well preserving garment texture and person identity.
For trousers, one can see that \emph{tapered} trousers gradually narrow toward the ankle, and \emph{straight} trousers maintain a uniform leg width.
These visualizations intuitively demonstrate that FitControler generates perceptually realistic and well-controlled fits for VTON.

\subsection{Ablation Study}

\subsubsection{Condition Injection.} 
\begin{figure*}[t]
  \centering
  \includegraphics[width=\linewidth]{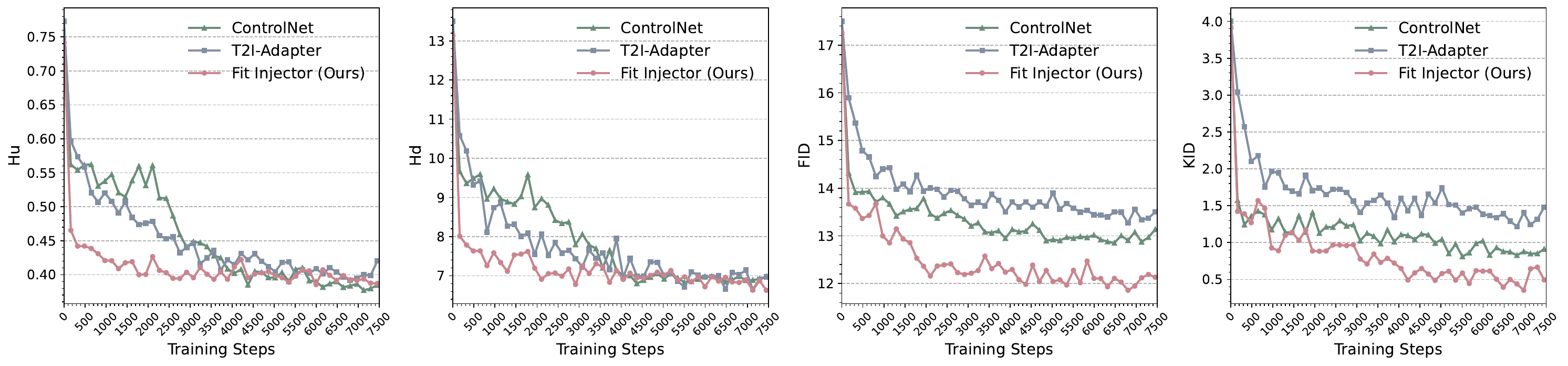}
  \vspace{-15pt}
  \caption{\textbf{Training curves of different conditional injection approaches.} Our approach converges faster with superior image quality.}
  \label{fig:ablation_control}
\end{figure*}

We compare our multi-scale fit injector with alternative controlled image generation approaches including ControlNet~\cite{zhang2023adding} and T2I-Adapter~\cite{mou2024t2i} on the short-sleeve T-shirts under the paired setting.
Table~\ref{tab:control} shows that our multi-scale fit injector achieves superior generation quality (FID/KID) while maintaining comparable control ability (Hu/Hd) to ControlNet, but with substantially fewer number of parameters. 
Fig.~\ref{fig:ablation_control} illustrates the training curves. Our multi-scale fit injector converges faster and consistently outperforms the baselines in terms of the image quality metrics (FID/KID) throughout the training process. Notably, it can achieve reasonable fit manipulation with only $1,000$ training steps.

\begin{table}[t!]
\centering
\scriptsize
% \caption{Ablation studies on FitControler. 
% Best performance is in \textbf{boldface}.}

% 上方：第一张和第三张表格左右排列
\begin{minipage}[t]{0.49\textwidth}
    \centering
    \setlength{\tabcolsep}{2pt}
    \renewcommand{\arraystretch}{1.3}
    \captionof{table}{\textbf{Condition injection}. Comparison of fit injector with alternative injection approaches after $7,500$ training steps. }
    \vspace{-10pt}
    % \resizebox{\textwidth}{!}{
    \begin{tabular}[t]{@{}lcccc|c}
    \toprule
    Method & FID & KID & Hu & Hd & \makecell{\tiny Trainable \\ \tiny Params} \\
    \midrule
    ControlNet & 13.14 & 0.91 & \textbf{0.38} & 6.98 & 361M \\
    T2I-Adapter & 13.50 & 1.48 & 0.42 & 6.97 & 77M \\
    Fit Injector & \textbf{12.13} & \textbf{0.49} & 0.39 & \textbf{6.63} & 97M \\
    \bottomrule
    \end{tabular}
    % }
    \label{tab:control}
\end{minipage}%
\hfill
\begin{minipage}[t]{0.49\textwidth}
    \centering
    \setlength{\tabcolsep}{6pt}
    \renewcommand{\arraystretch}{1.17}
    \captionof{table}{\textbf{Training sample size.} FitControler trained on 1,000 samples matches full-data performance.}
        \vspace{-10pt}
    % \resizebox{\textwidth}{!}{
    \begin{tabular}[t]{@{}lcccc}
    \toprule
    \# Samples & FID & KID & Hu & Hd \\
    \midrule
    1000 & 17.50 & 1.56 & 0.44 & 8.39 \\
    2000 & 17.51 & 1.42 & 0.44 & \textbf{7.21} \\
    3000 & \textbf{17.46} & \textbf{1.31} & \textbf{0.44} & 8.12 \\
    3959 (Full) & 17.80 & 1.52 & 0.45 & 8.25 \\
    \bottomrule
    \end{tabular}
    % }
    \label{tab:train_size}
\end{minipage}
\end{table}

% % 下方：第二张表格单独居中
% \begin{table}[t]
%     \centering
%     \scriptsize
%     \setlength{\tabcolsep}{4pt}
%     \renewcommand{\arraystretch}{1.2}
%     \captionof{table}{\textbf{FitControler vs. extra training}. FitControler accounts for major improvements on Hu and Hd, while extra training (finetune) yields only marginal gains.}
%     \vspace{-10pt}
%     % \resizebox{0.8\textwidth}{!}{ % 可以根据需要调整宽度比例，这里用0.8\textwidth
%     \begin{tabular}[t]{@{}lcccccc}
%     \toprule
%     Method & FitCtrl & Finetune & FID & KID & Hu & Hd \\
%     \midrule
%     \multirow{3}*{Leffa}& \ding{55} &\ding{55} &18.77 & 2.59& 0.65&10.47 \\
%                          & \ding{55} &\ding{51} & 18.69 & 2.18 & 0.63 & 12.67 \\
%                          & \ding{51} &\ding{55} &\textbf{17.73} & \textbf{1.54}& \textbf{0.45}&\textbf{7.65} \\
%     \midrule
%     \multirow{3}*{FitDiT}& \ding{55} &\ding{55} &23.94& 6.82& 0.80& 16.89 \\
%                          & \ding{55} &\ding{51} & 19.49 & 3.55 & 0.81 & 14.42 \\
%                          & \ding{51} &\ding{55} &\textbf{18.48} & \textbf{1.86}& \textbf{0.46}&\textbf{7.94} \\
%     \bottomrule
%     \end{tabular}

% \label{tab:ablation_trainset}
% \end{table}

\vspace{-10pt}
\subsubsection{Training Sample Size.}
Here we study the impact of training sample size on FitControler by training both the layout generator and fit injector with Leffa~\cite{zhou2025learning} on the short-sleeve data of varying sample sizes ($1,000$, $2,000$, $3,000$, and $3,959$ samples) and test under the unpaired setting. According to Table~\ref{tab:train_size}, FitControler achieves good performance even with only $1,000$ training samples, implying high sample efficiency.

\vspace{-10pt}
\subsubsection{FitControler vs. Extra Training.} To verify that the observed performance mainly originates from FitControler rather than additional training on Fit4Men, here we conduct an ablation study on short-sleeve shirts. We compare three configurations:
(1) \emph{Baseline}: the original VTON model without both FitControler and finetuning;
(2) \emph{Backbone finetuning}: finetuning the VTON model on Fit4Men using our garment-agnostic preprocessor, but without FitControler;
(3) \emph{FitControler only}: inserting FitControler while freezing the original VTON model.
As shown in Table~\ref{tab:ablation_trainset}, backbone finetuning slightly improves the FID and KID but has negligible effect on fit control. 
In contrast, incorporating FitControler significantly reduces Hu and Hd, indicating better adherence to the target fit. 
These results suggest that the performance gains indeed come from FitControler. 
In contrast, extra training only contributes to minor image quality improvements in this context.

\begin{table}[t!]
\centering
\scriptsize
% \caption{Ablation studies on FitControler. 
% Best performance is in \textbf{boldface}.}

% 上方：第一张和第三张表格左右排列
\begin{minipage}[t]{0.39\textwidth}
    \centering
    \scriptsize
    \setlength{\tabcolsep}{1pt}
    \renewcommand{\arraystretch}{1.37}
    \captionof{table}{\textbf{FitControler vs. extra training}. FitControler accounts for major improvements on Hu and Hd, while extra finetuning yields only marginal gains.}
    \vspace{-10pt}
    \resizebox{\textwidth}{!}{ % 可以根据需要调整宽度比例，这里用0.8\textwidth
    \begin{tabular}[t]{@{}lcccccc}
    \toprule
    Method & FitCtrl & Finetune & FID & KID & Hu & Hd \\
    \midrule
    \multirow{3}*{Leffa}& \ding{55} &\ding{55} &18.77 & 2.59& 0.65&10.47 \\
                         & \ding{55} &\ding{51} & 18.69 & 2.18 & 0.63 & 12.67 \\
                         & \ding{51} &\ding{55} &\textbf{17.73} & \textbf{1.54}& \textbf{0.45}&\textbf{7.65} \\
    \midrule
    \multirow{3}*{FitDiT}& \ding{55} &\ding{55} &23.94& 6.82& 0.80& 16.89 \\
                         & \ding{55} &\ding{51} & 19.49 & 3.55 & 0.81 & 14.42 \\
                         & \ding{51} &\ding{55} &\textbf{18.48} & \textbf{1.86}& \textbf{0.46}&\textbf{7.94} \\
    \bottomrule
    \end{tabular}

    }
\label{tab:ablation_trainset}
\end{minipage}%
\hfill
\begin{minipage}[t]{0.59\textwidth}
    \centering
    \addtolength{\tabcolsep}{1pt}
    \renewcommand{\arraystretch}{1.1}
    \captionof{table}{\textbf{FitControler vs. prompt conditioning.} While prompt conditioning improves fit control to some extent, FitControler consistently yields superior fit consistency (Hu/Hd) and overall image quality.}
        \vspace{-6pt}
    \resizebox{\linewidth}{!}{
    \begin{tabular}{lcccccccccc}
    \toprule
    \multirow{2}*{Method} & \multicolumn{6}{c}{Paired} & \multicolumn{4}{c}{Unpaired}\\
    \cmidrule(r){2-7}\cmidrule(r){8-11}
    &FID$\downarrow$ & KID$\downarrow$ & SSIM$\uparrow$ & LPIPS$\downarrow$ & Hu$\downarrow$ & Hd$\downarrow$
    &FID$\downarrow$ & KID$\downarrow$ & Hu$\downarrow$ & Hd$\downarrow$ \\
    \midrule
    \textbf{Short-sleeve} \\
    IDMVTON
    &15.99 &2.90 &0.852 &0.0647 &0.55 &9.03
    &20.83 &3.37 &0.67 &10.39 \\
    +Prompt
    &15.49 &1.98 &\textbf{0.868} &\textbf{0.0581} &0.46 &8.38
    &21.28 &3.29 &0.56 &9.77 \\
    +FitControler
    &\textbf{14.69} 
    &\textbf{1.92}
    &0.866
    &0.0592 
    &\textbf{0.45} 
    &\textbf{7.54} 
    &\textbf{19.11}
    &\textbf{2.19} 
    &\textbf{0.51} 
    &\textbf{8.81} \\
    
    \midrule
    \textbf{Trousers} \\
    IDMVTON
    &15.11 &6.67 &0.854 &0.0677 &1.03 &10.53
    &16.12 &5.59 &1.58 &12.58 \\
    +Prompt
    &13.07 &4.14 &0.860 &\textbf{0.0577} &0.89 &9.68
    &15.11 &4.96 &1.32 &11.38 \\
    +FitControler
    &\textbf{12.14} 
    &\textbf{3.68}
    &\textbf{0.862}
    &0.0592
    &\textbf{0.81}
    &\textbf{9.32}
    &\textbf{14.43} 
    &\textbf{3.95}
    &\textbf{1.13}
    &\textbf{10.29}\\
    
    \bottomrule
    \end{tabular}
    }
    \label{tab:compare_prompt}
\end{minipage}
\end{table}

% \begin{minipage}[t]{0.49\textwidth}
% \captionsetup{type=figure}
%     \centering
%         \includegraphics[width=\linewidth]{images/vs_prompt.pdf}
%         \captionof{figure}{Qualitative comparison of IDMVTON: text control vs. FitControler.}
%         \label{fig:vs_prompt}
% \end{minipage}%
% \begin{figure}[t!]
%     \centering
%     \includegraphics[width=\linewidth]{images/vs_prompt.pdf}
%     \caption{Qualitative comparison of IDMVTON: text control vs. FitControler.}
%     \label{fig:vs_prompt}
% \end{figure}

\vspace{-10pt}
\subsubsection{FitControler vs. Prompt Conditioning.} 
\begin{figure}[t!]
    \centering
    \begin{minipage}{0.49\linewidth}
        \centering
        \includegraphics[width=\linewidth]{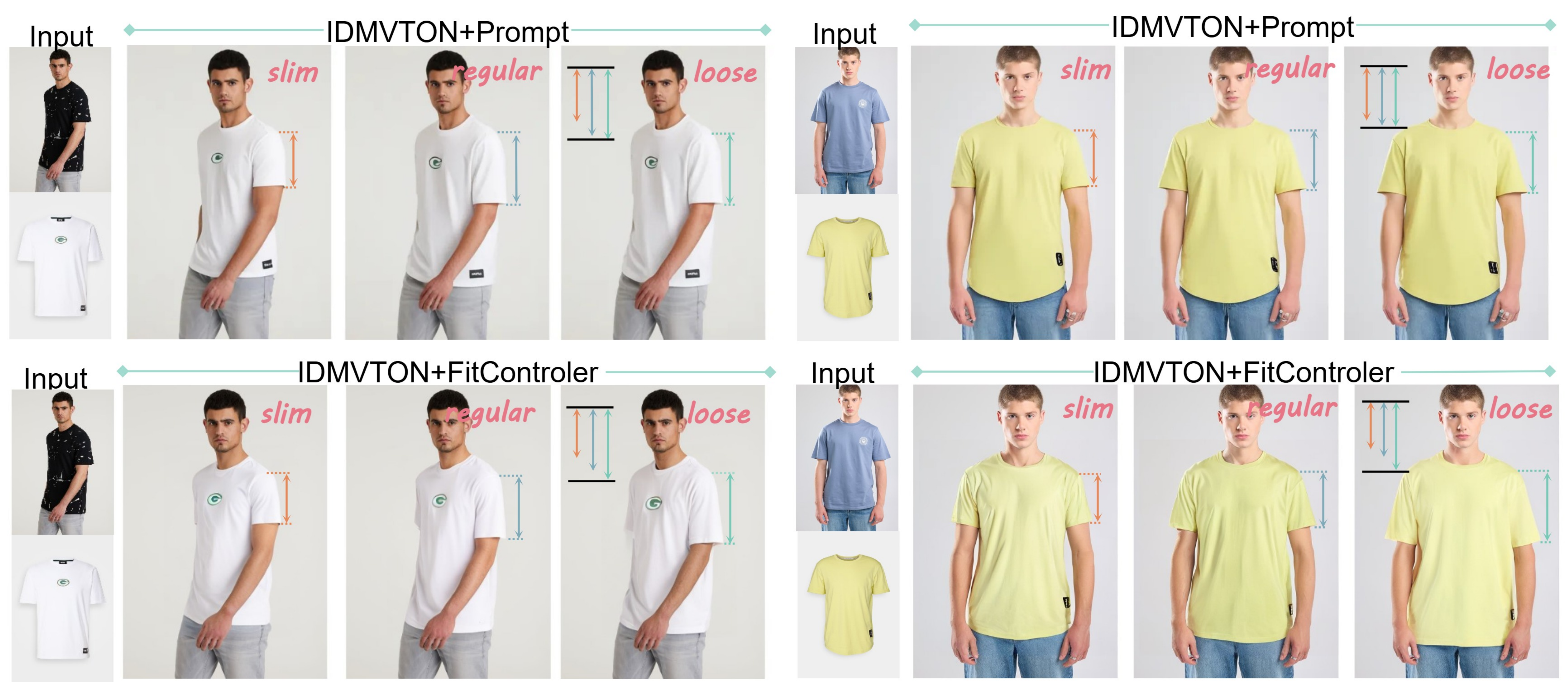}
        \caption{Qualitative comparison of IDMVTON: text control vs. FitControler.}
        \label{fig:vs_prompt}
    \end{minipage}
    \hfill
    \begin{minipage}{0.49\linewidth}
    \vspace{0pt}
        \centering
        \includegraphics[width=\linewidth]{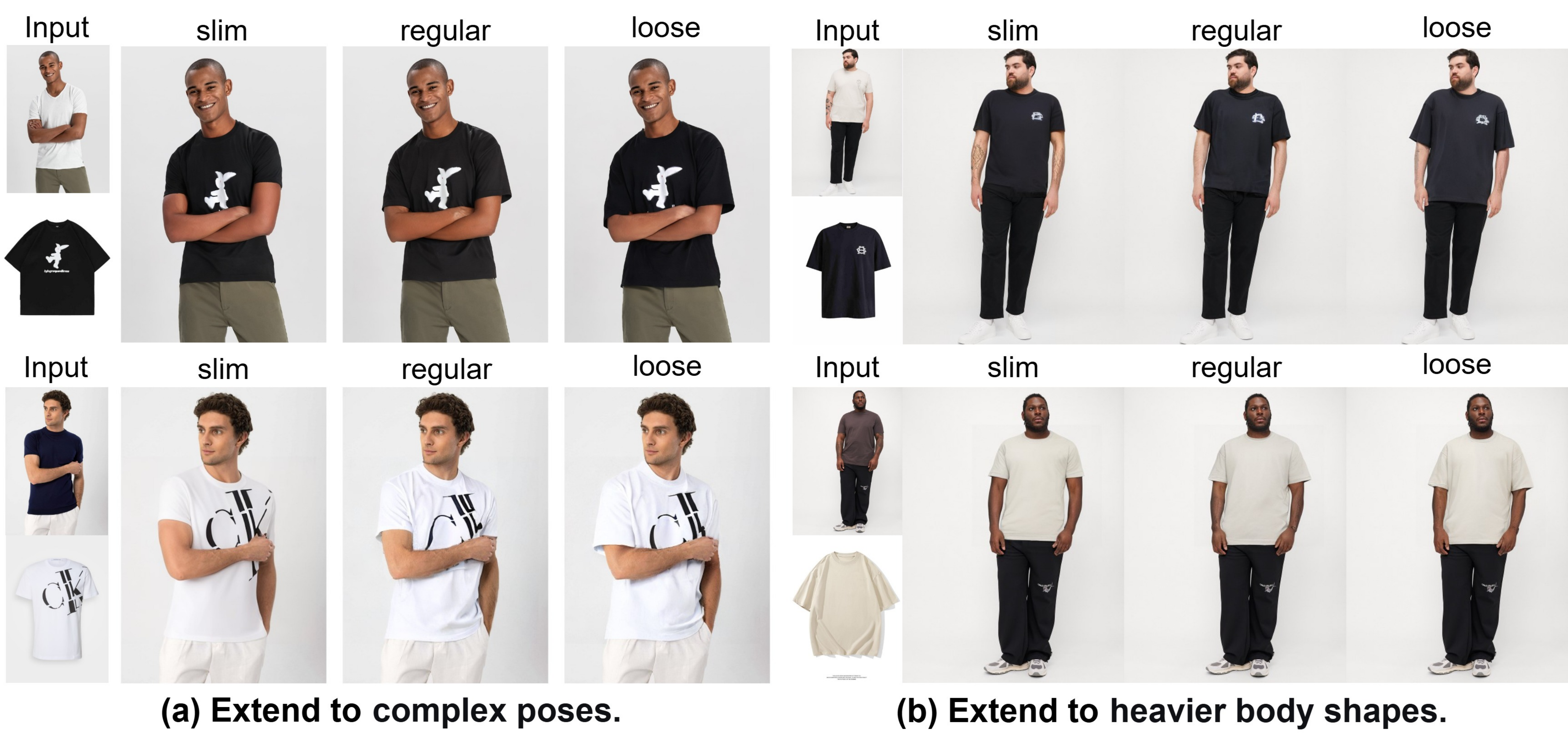}
        \vspace{-18pt}
        \caption{Results of FitControler extended to diverse poses and body shapes.}
        \label{fig:complex_pose}
    \end{minipage}
\end{figure}

To investigate whether fit control can be achieved by text conditioning alone, we construct an IDM-VTON baseline by appending fit labels to the text prompts and fine-tuning the model for $16{,}000$ steps. We find that prompt conditioning alone is ineffective under standard preprocessing due to shape leakage, and only becomes viable when combined with our garment-agnostic representation. As shown in Table~\ref{tab:compare_prompt}, although prompt conditioning improves fit consistency over the vanilla IDM-VTON, it consistently underperforms FitControler across both short-sleeve and trouser scenarios. Qualitative comparisons in Fig.~\ref{fig:vs_prompt} further show that prompt-based control yields weaker fit magnitude and less accurate garment boundaries. These results demonstrate that text conditioning alone is insufficient for precise fit control, while explicit spatial modeling in FitControler provides substantially better fit-aware generation.

% \begin{wrapfigure}{r}{0.5\linewidth}
% \centering  
% \begin{center}
% \vspace{-35pt}
% \includegraphics[width=\linewidth]{images/complex_pose.pdf}
% \end{center}
% \vspace{-10pt}
% \caption{Results of FitControler extended to complex poses, heavier body shapes.}
% \label{fig:complex_pose}  
% \vspace{-20pt}
% \end{wrapfigure}

\subsection{Extension to Diverse Poses and Body Shapes}
To further evaluate the generalization ability of FitControler, we additionally test it on models with diverse body shapes and poses. As shown in Fig.~\ref{fig:complex_pose}(b) and supplementary Fig.~S4, we directly use the original DensePose representation to preserve body-shape information. Despite minor body-shape preservation errors (e.g., slightly thinner appearances under slim-fit settings), our method consistently achieves the desired fit control across different body types. Furthermore, the complex-pose examples in Fig.~\ref{fig:complex_pose}(a) demonstrate that our framework maintains stable fit control under challenging pose variations. These results suggest that FitControler generalizes beyond the controlled training setting, while more diverse body-shape distributions remain an important direction for future study.

\section{Conclusion}

We introduced FitControler, a novel plug-in module that equips diffusion-based VTON models with precise control over garment fit. It first models garment layouts under specified fit through a fit-aware layout generator conditioned on carefully designed garment-agnostic representations, and then conveys these layout features to VTON models via a multi-scale fit injector, enabling layout-driven generation.
In particular, we built a dedicated fit-aware dataset, Fit4Men, and proposed two evaluation metrics specifically tailored to assess fit control. Extensive experiments demonstrate that our approach effectively bridges the gap between high-level fit concepts and realistic try-on image synthesis, providing a practical solution for controllable virtual try-on. 

Our current dataset is restricted to menswear, covering only a limited range of garment categories and fit types, and does not sufficiently account for diverse body shapes. These factors may inevitably constrain the diversity and generalization capability of the proposed framework.
For future work, we plan to expand the dataset to include womenswear, a broader spectrum of garment categories, richer fit variations, and models with diverse body types, thereby further enhancing the applicability and robustness of fit-aware virtual try-on systems in real-world scenarios.

% \clearpage  % TODO FINAL: This \clearpage needs to be removed from both review and camera-ready versions.

\section*{Acknowledgements}
This work is supported by the National Natural Science Foundation of China under Grant No. 62576146.

\bibliographystyle{splncs04}
\bibliography{main}

\clearpage
\input{supplementary/X_suppl}
\end{document}

%% file: supplementary/X_suppl.tex
% \clearpage
% \setcounter{page}{1}
\appendix
\begin{center}
\vspace{20pt}
{\bf {\LARGE Appendix}}
\vspace{15pt}
\end{center}

\setcounter{table}{0}

\setcounter{figure}{0}\setcounter{section}{0}
\setcounter{equation}{0}
\renewcommand{\thetable}{S\arabic{table}}
\renewcommand{\thefigure}{S\arabic{figure}}
\renewcommand{\thesection}{S\arabic{section}}
\renewcommand{\theequation}{S\arabic{equation}}

% \begin{figure}
%   \centering
%    \includegraphics[width=\linewidth]{supplementary/details_for_layout_generator.pdf}
%    \caption{\textbf{Overview of Layout Generator.} The layout generator predicts the layout image conditioned on the fit label and is trained with cross-entropy loss. }
%    %The layout generator is responsible for predicting the layout image conditioned on the fit label, and is trained using the cross-entropy loss.
%    \label{fig:details_for_layout_generator}
% \end{figure}

The supplementary material is organized into five sections. 
Section~\ref{sec:Details about layout generator} describes the fit-aware layout generator in detail. 
Section~\ref{sec:Training Details} provides additional training information for both the layout generator and the fit injector.  
Section~\ref{sec:additional ablation} reports extended ablation studies on FitControler.
Section~\ref{sec:additional results} presents further experimental results.
Finally, Section~\ref{sec:limitations} discusses the limitations of FitControler and outlines potential directions for future research.

\begin{wrapfigure}{r}{0.6\linewidth}
\centering  
\begin{center}
\vspace{-35pt}
\includegraphics[width=\linewidth]{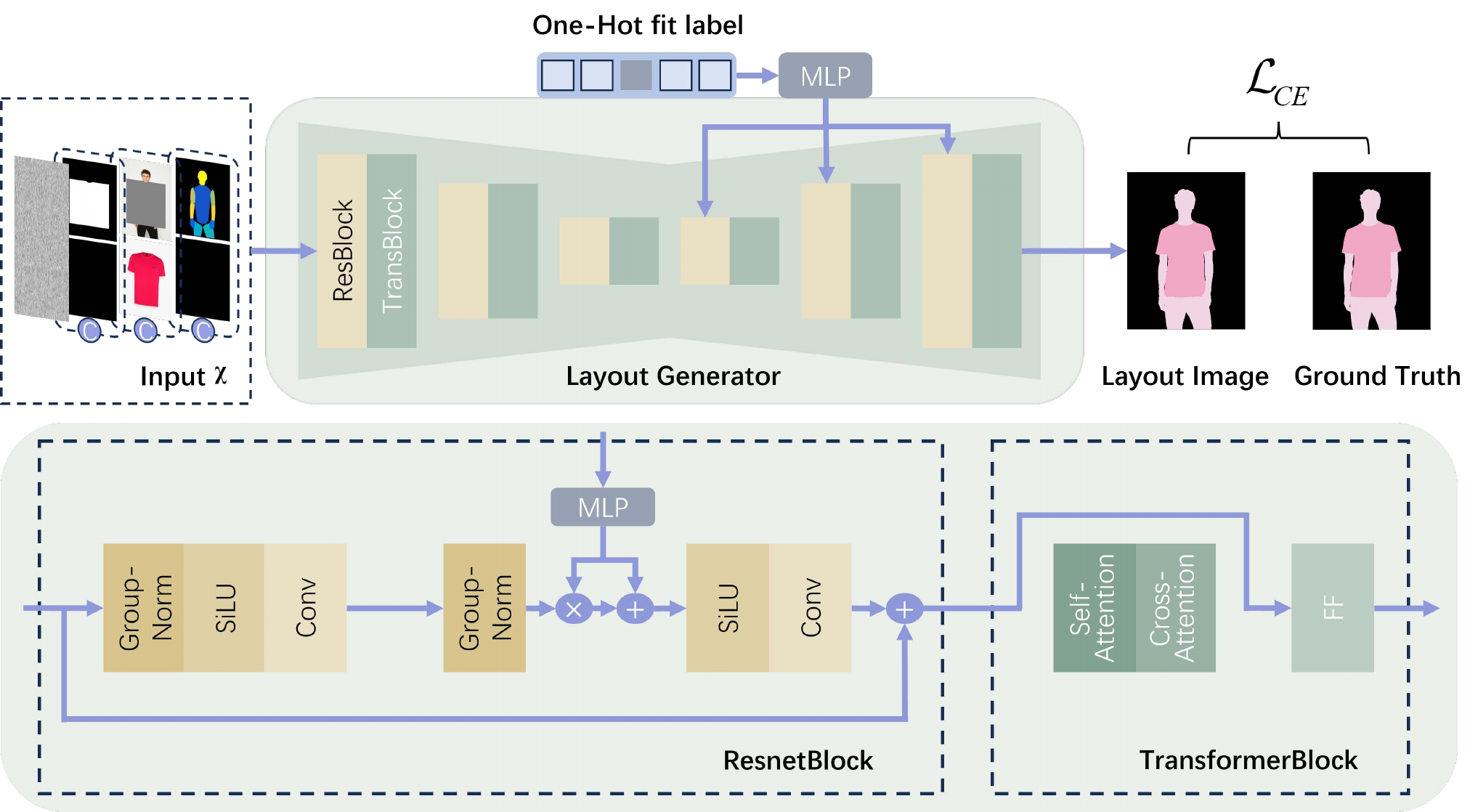}
\end{center}
\vspace{-10pt}
\caption{\textbf{Overview of Layout Generator.} The layout generator predicts the layout image conditioned on the fit label and is trained with cross-entropy loss. }
\label{fig:details_for_layout_generator}  
\vspace{-20pt}
\end{wrapfigure}

\section{Details for Fit-Aware Layout Generator}
\label{sec:Details about layout generator}

 As shown in Fig.~\ref{fig:details_for_layout_generator}, we repurpose the U-Net from the Stable Diffusion v1.5 inpainting version~\cite{rombach2022high} as our layout generator. The U-Net takes a 13-channel tensor $\mathcal{X}$ and a one-hot encoded fit label $\bm{l}$ as input, and generates a 3-channel segmentation map $\bm{S}_l$ along with fit-aware features $\bm{C}_l$. To accommodate our input–output design, we extend its initial convolution layer to 13 channels and modify the output layer to 3 channels. To incorporate $\bm{l}$ as a control signal, we apply FiLM~\cite{perez2018film} modulation in each ResNet block to inject the fit information into the decoder features and remove all attention operations from the decoder. Moreover, we extract the features preceding each residual block in the decoder and aggregate them as multi-scale features $\bm{C}_l$. The encoder remains frozen to preserve its semantic representation capability.
 
\section{Training Details}
\label{sec:Training Details}
\subsubsection{Layout Generator.}
We expand input channels of the U-Net from 9 to 13 by adding four additional channels for dense pose, which disrupts the pretrained semantic priors in the original U-Net. To recover these semantics, we first modify only the input layer and train a try-on model for 16,000 steps on the VITON-HD dataset following the CatVTON~\cite{chong2024catvton} configuration to learn pose encoding.

Next, we freeze the encoder and modify the decoder (without FiLM) and the output layer as described in Sec.~\ref{sec:Details about layout generator}. We train the model for 10,000 steps on the VITON-HD~\cite{choi2021viton} and DressCode~\cite{morelli2022dress} datasets for fit-unaware layout prediction.

Finally, we introduce FiLM modulation and fine-tune the decoder on the Fit4Men dataset for 10,000 steps in a fit-aware setting. Both short-sleeve tops and trousers are trained within the same model. 

\vspace{-10pt}
\subsubsection{Fit Injector.}
When applied to different VTON models, we keep the layout generator fixed and train a separate fit injector for each model. For all VTON models as mentioned in Sec. 5.1, the fit injector is trained for 7,500 steps using the MSE loss.

Throughout all stages, we use AdamW~\cite{loshchilov2017decoupled} with a learning rate of $1\times10^{-5}$ and a batch size of 64.

\begin{wrapfigure}{r}{0.5\textwidth} % {r}表示右侧，宽度0.5\textwidth
    \centering
    \vspace{-25pt}
    % 上方图片
    \includegraphics[width=\linewidth]{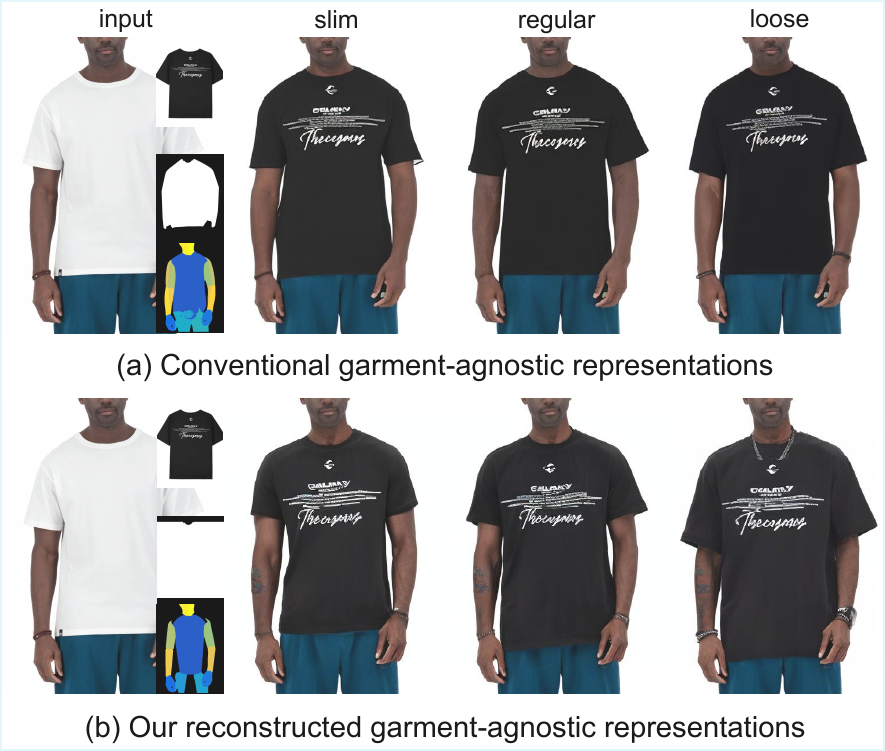}
    \caption{\textbf{Ablation study on the garment-agnostic preprocessor.} 
    (a) FitControler trained with commonly used garment-agnostic representations. 
    (b) FitControler trained with our proposed representations. 
    Our method enables more faithful fit-aware control and better garment adaptation.}
    \label{fig:ablation_processor}
    
    % \vspace{1em} % 图片和表格之间的间距
    
    % 下方表格
    \centering
    \footnotesize
    \setlength{\tabcolsep}{2pt}
    \captionof{table}{Ablation on semantic priors for the layout generator.
    Initializing the layout generator with pretrained diffusion weights (semantic prior) yields better performance.}
    \renewcommand{\arraystretch}{1.2}
    \resizebox{0.5\textwidth}{!}{
    \begin{tabular}{@{}l|cccc}
    \toprule
    Semantic Prior & FID $\downarrow$ & KID $\downarrow$ & Hu $\downarrow$ & Hd $\downarrow$ \\
    \midrule
    Without Prior (Random Init.) & 13.35 & 1.11 & 0.41 & 7.91 \\
    With Diffusion Prior & \textbf{12.45} & \textbf{0.62} & \textbf{0.40} & \textbf{7.05} \\
    \bottomrule
    \end{tabular}
    }
    \label{tab:ablation_seg}
    \vspace{-20pt}

\end{wrapfigure}

\section{Ablation Study}
\label{sec:additional ablation}
\subsubsection{Garment-Agnostic Preprocessor.} 
Fig.~\ref{fig:ablation_processor} compares the results of FitControler trained with commonly used garment-agnostic representations versus our proposed ones. 
When trained with the conventional representations, FitControler tends to preserve the original garment shape and fails to adjust the fit according to the given label. 
In contrast, our garment-agnostic preprocessor allows FitControler to generate images that accurately reflect the specified fit conditions. 
These results demonstrate the effectiveness of our design and its suitability for fit-aware VTON.

\vspace{-10pt}
\subsubsection{Effect of Semantic Priors for the Layout Generator.}
We examine whether incorporating semantic priors benefits the layout generator.
Without semantic priors, the layout generator is randomly initialized; 
with them, it is initialized using pretrained diffusion weights.
As shown in Table~\ref{tab:ablation_seg}, the diffusion-based initialization improves all metrics, indicating that semantic knowledge from pretrained diffusion models facilitates more accurate body–garment segmentation and enhances performance.

\begin{table*}[t]
\centering
\footnotesize
\renewcommand{\arraystretch}{1.1}
\addtolength{\tabcolsep}{1pt}
\newcommand{\imp}[1]{\raisebox{-0ex}{\hspace{1pt}\scriptsize\textcolor{ForestGreen}{#1}}} % improvement (下降指标)
\newcommand{\impup}[1]{\raisebox{-0ex}{\hspace{1pt}\scriptsize\textcolor{NiceRed}{#1}}} % 上升指标（红色）
\caption{Performance of VTON models with/without FitControler on regular-fit subset. Relative improvements are colored in \textcolor{NiceRed}{red} $\uparrow$ and \textcolor{ForestGreen}{green} $\downarrow$.}
\vspace{-5pt}
% \begin{tabular}{lcccccccccc}
\resizebox{\textwidth}{!}{
\begin{tabular}{lllllllllll}
\toprule
\multirow{2}*{Method} & \multicolumn{6}{c}{Paired} & \multicolumn{4}{c}{Unpaired}\\
\cmidrule(r){2-7}\cmidrule(r){8-11}
&FID$\downarrow$ & KID$\downarrow$ & SSIM$\uparrow$ & LPIPS$\downarrow$ & Hu$\downarrow$ & Hd$\downarrow$
&FID$\downarrow$ & KID$\downarrow$ & Hu$\downarrow$ & Hd$\downarrow$ \\
\midrule
FitDiT~\cite{jiang2024fitdit}
&24.97 &4.69 &0.873 &0.0627 &0.63 &11.51
&39.21 &12.15 &0.76 &15.40 \\
\rowcolor{gray!10}+FitControler
&\textbf{18.17}\imp{(27.2\%)} 
&\textbf{0.57}\imp{(87.8\%)} 
&\textbf{0.887}\impup{(1.6\%)} 
&\textbf{0.0445}\imp{(29.0\%)} 
&\textbf{0.36}\imp{(42.8\%)} 
&\textbf{6.76}\imp{(41.3\%)} 
&\textbf{29.76}\imp{(24.1\%)} 
&\textbf{3.49}\imp{(71.2\%)} 
&\textbf{0.44}\imp{(42.1\%)} 
&\textbf{8.04}\imp{(47.8\%)} \\
Leffa~\cite{zhou2025learning}
&20.44 &1.22 &0.876 &0.0540 &0.50 &7.34
&28.15 &3.73 &0.53 &7.84 \\
\rowcolor{gray!10}+FitControler
&\textbf{18.07}\imp{(11.6\%)} 
&\textbf{0.23}\imp{(81.1\%)} 
&\textbf{0.882}\impup{(0.6\%)} 
&\textbf{0.0456}\imp{(15.6\%)} 
&\textbf{0.39}\imp{(22.0\%)} 
&\textbf{6.66}\imp{(9.3\%)} 
&\textbf{27.83}\imp{(1.1\%)} 
&\textbf{2.61}\imp{(30.0\%)} 
&\textbf{0.46}\imp{(13.2\%)} 
&\textbf{7.08}\imp{(9.7\%)} \\

\bottomrule
\end{tabular}
}
\label{tab:sup_compare}
\end{table*}

\begin{figure}[t]
    \centering
    \includegraphics[width=\linewidth]{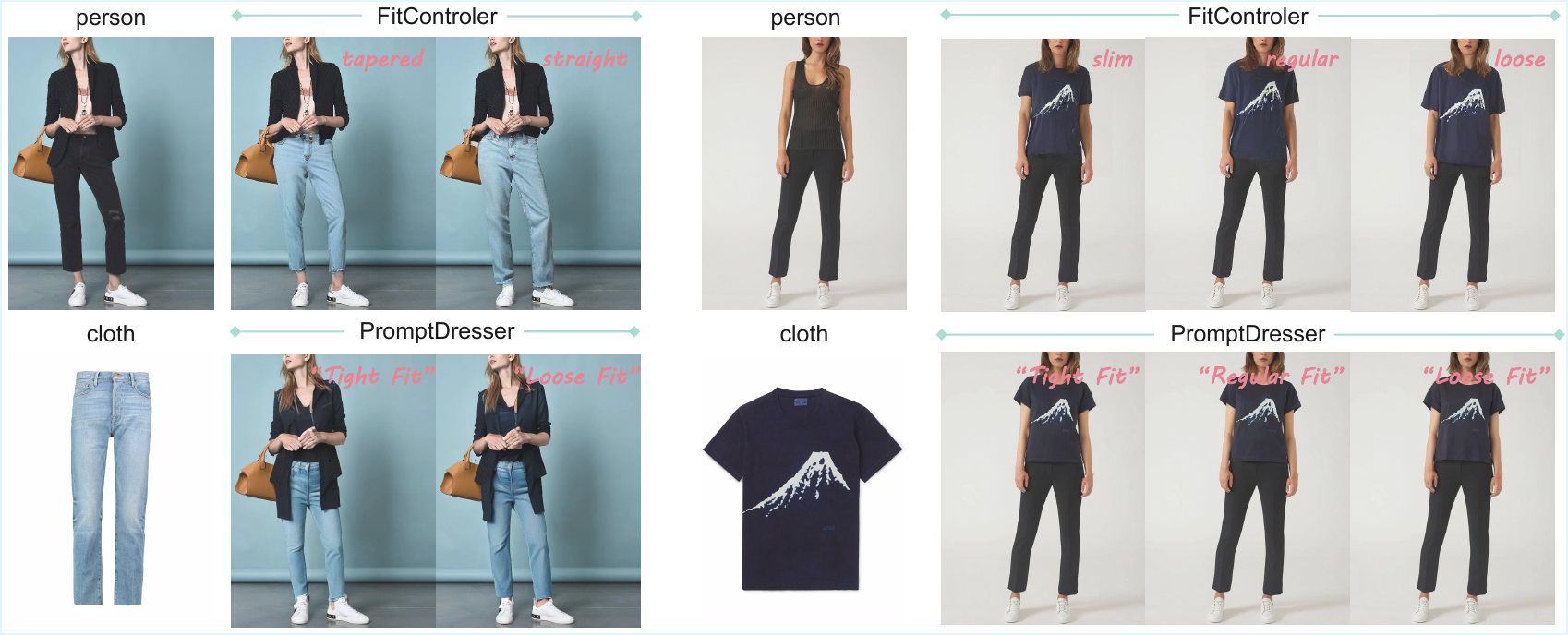}
    \caption{\textbf{Comparison with PromptDresser on DressCode.}
    PromptDresser exhibits limited sensitivity to fit-related textual descriptions, whereas FitControler produces results that more faithfully reflect the specified fit conditions.}
    \label{fig:promptdresser}
\end{figure}

\section{Additional Experimental Results}
\label{sec:additional results}

\subsubsection{Comparison under the Regular-Fit Setting}
We observe that, for short-sleeve shirts, existing VTON models often generate regular-fit results by default. 
To examine whether FitControler still provides improvements under this tendency, we conduct a separate evaluation on samples labeled as regular fit, computing the corresponding metrics within this subset. 
This experiment complements the main results reported in Table~2.
As shown in Table~\ref{tab:sup_compare}, FitControler consistently outperforms the baseline models even when the target fit is regular. 
The improvements are comparable to those observed in the main experiments, indicating that the performance gains are not limited to more distinct fit variations (e.g., \textit{loose} or \textit{slim}), but also extend to the commonly generated regular-fit setting. 
These results further support the effectiveness and robustness of the proposed fit-aware control mechanism.

\vspace{-10pt}
\subsubsection{Comparison with PromptDresser~\cite{kim2024promptdresser}.}
We conduct a qualitative comparison with PromptDresser on the DressCode~\cite{morelli2022dress} dataset. 
The textual prompts for PromptDresser follow the template provided in its official implementation. 
As shown in Fig.~\ref{fig:promptdresser}, PromptDresser demonstrates limited adaptability to fit-related textual descriptions, while FitControler generates results that more consistently align with the intended fit specifications. 
This difference likely arises from PromptDresser’s reliance on implicit text-to-image associations, without explicit modeling of body--garment spatial relationships required for fit control.

\vspace{-10pt}
\subsubsection{FitControler with Other VTON Models.}
\begin{figure*}[t]
  \centering
   \includegraphics[width=\linewidth]{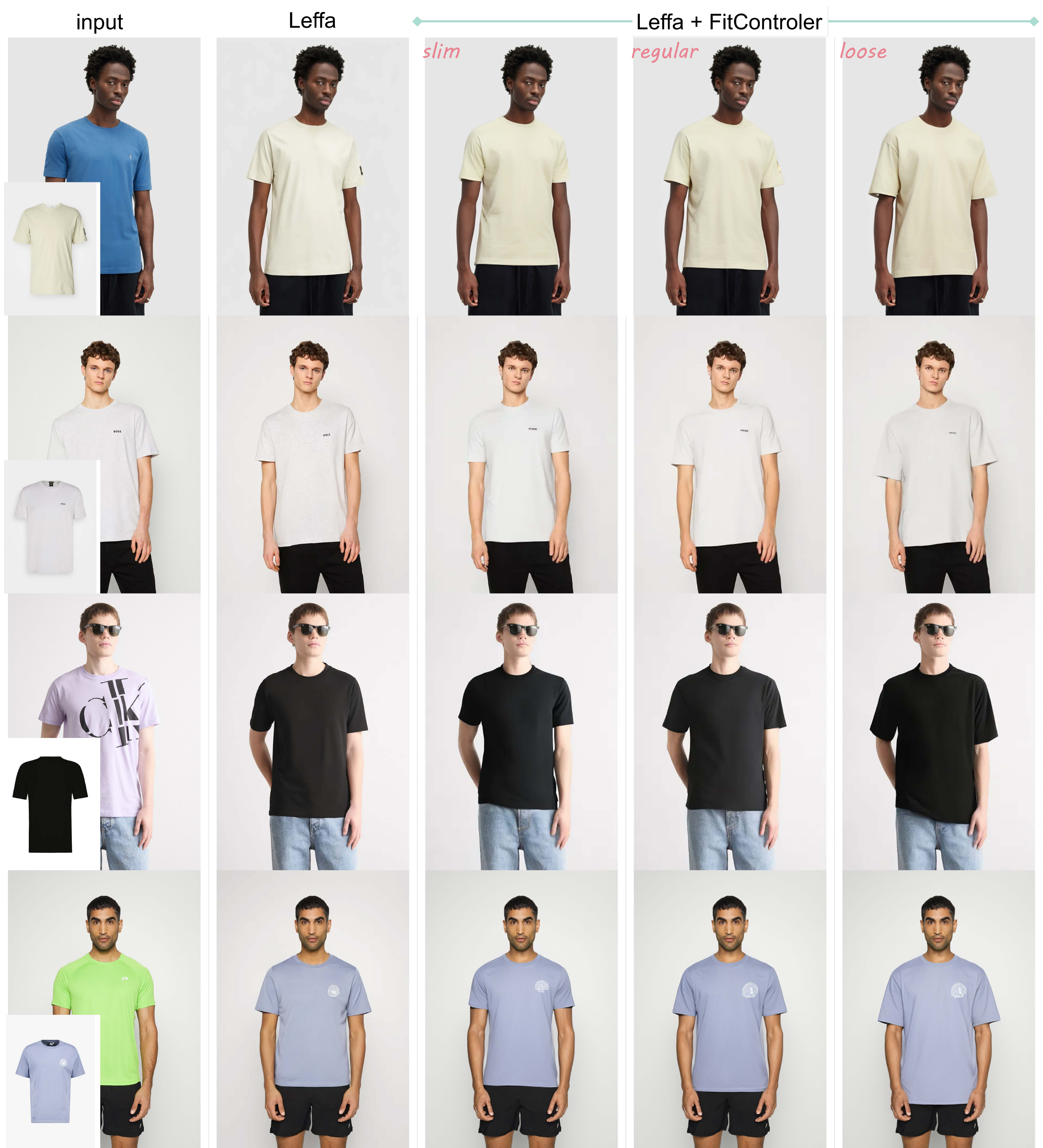}
   \caption{\textbf{Sample generated by Leffa without and with FitControler on short-sleeve shirts.}
   FitControler provides reliable fit control and maintains high visual fidelity for a wide range of poses.}
   \label{fig:onecol1}
\end{figure*}

\begin{figure*}[t]
  \centering
   \includegraphics[width=\linewidth]{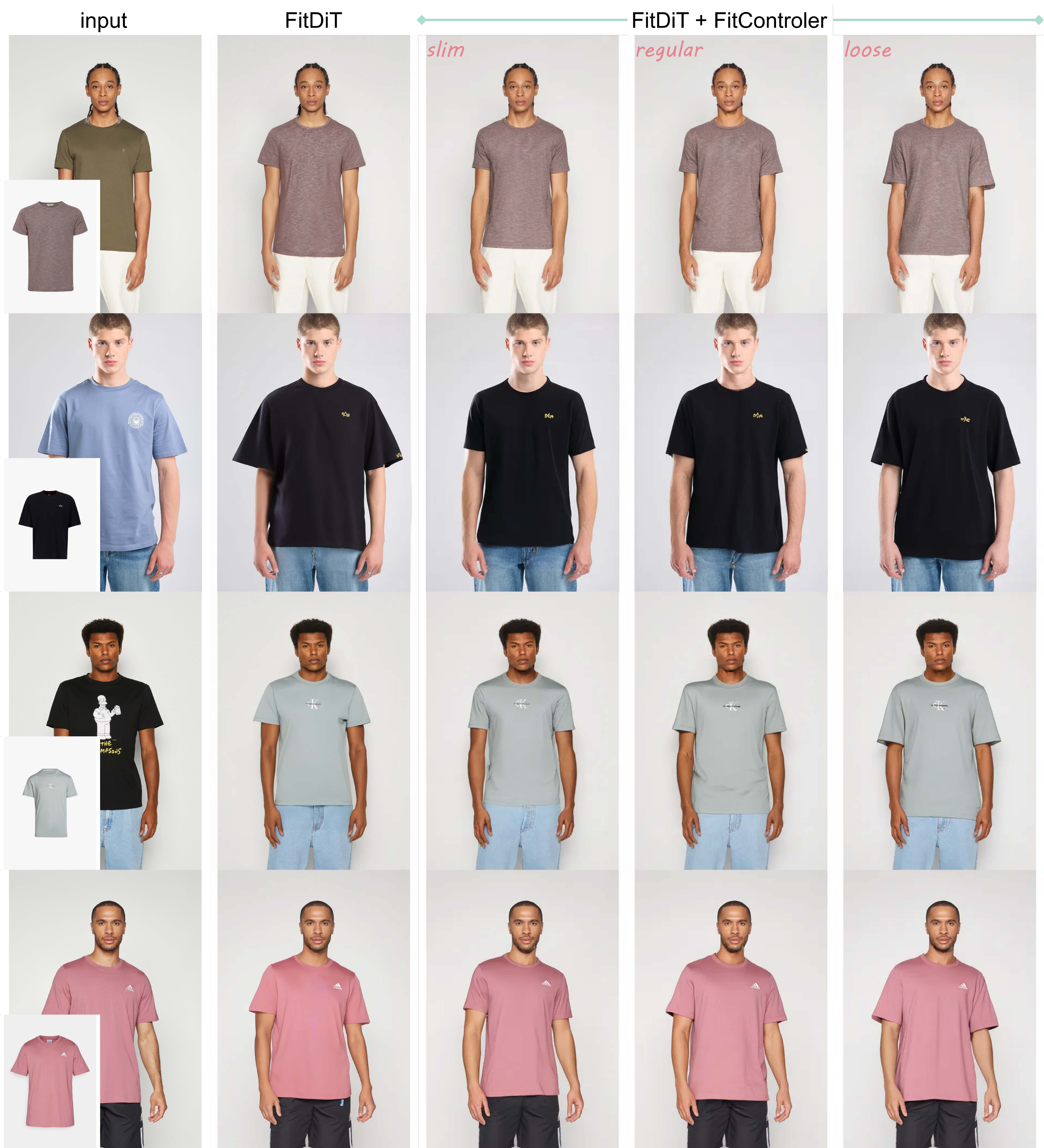}
   \caption{\textbf{Sample generated by FitDiT without and with FitControler on short-sleeve shirts.}
   FitControler provides reliable fit control and maintains high visual fidelity for a wide range of poses.}
   \label{fig:onecol2}
\end{figure*}

\begin{figure*}[t]
  \centering
   \includegraphics[width=\linewidth]{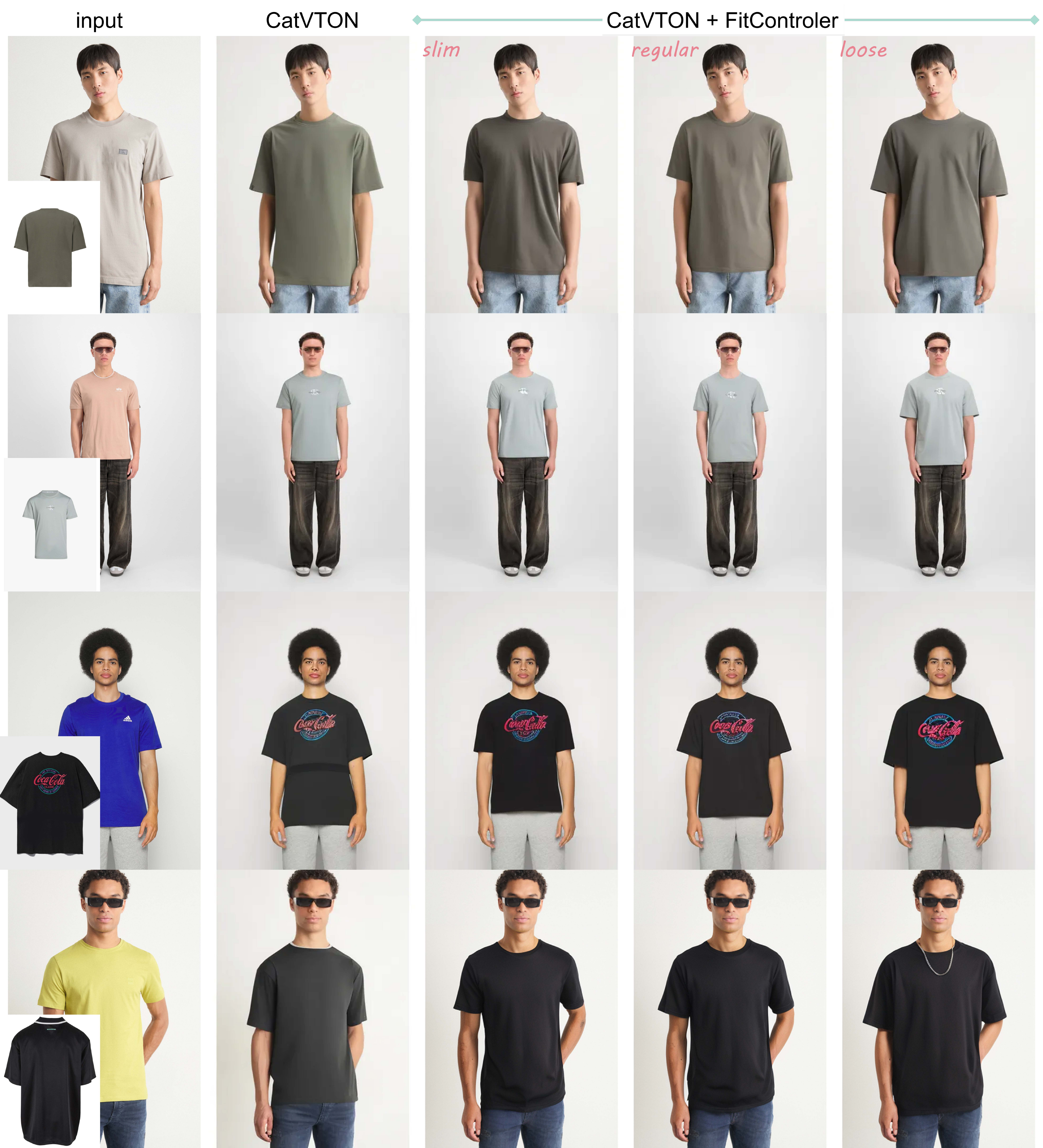}
   \caption{\textbf{Sample generated by CatVTON without and with FitControler on short-sleeve shirts.}
   FitControler provides reliable fit control and maintains high visual fidelity for a wide range of poses.}
   \label{fig:onecol3}
\end{figure*}

\begin{figure*}[t]
  \centering
   \includegraphics[width=\linewidth]{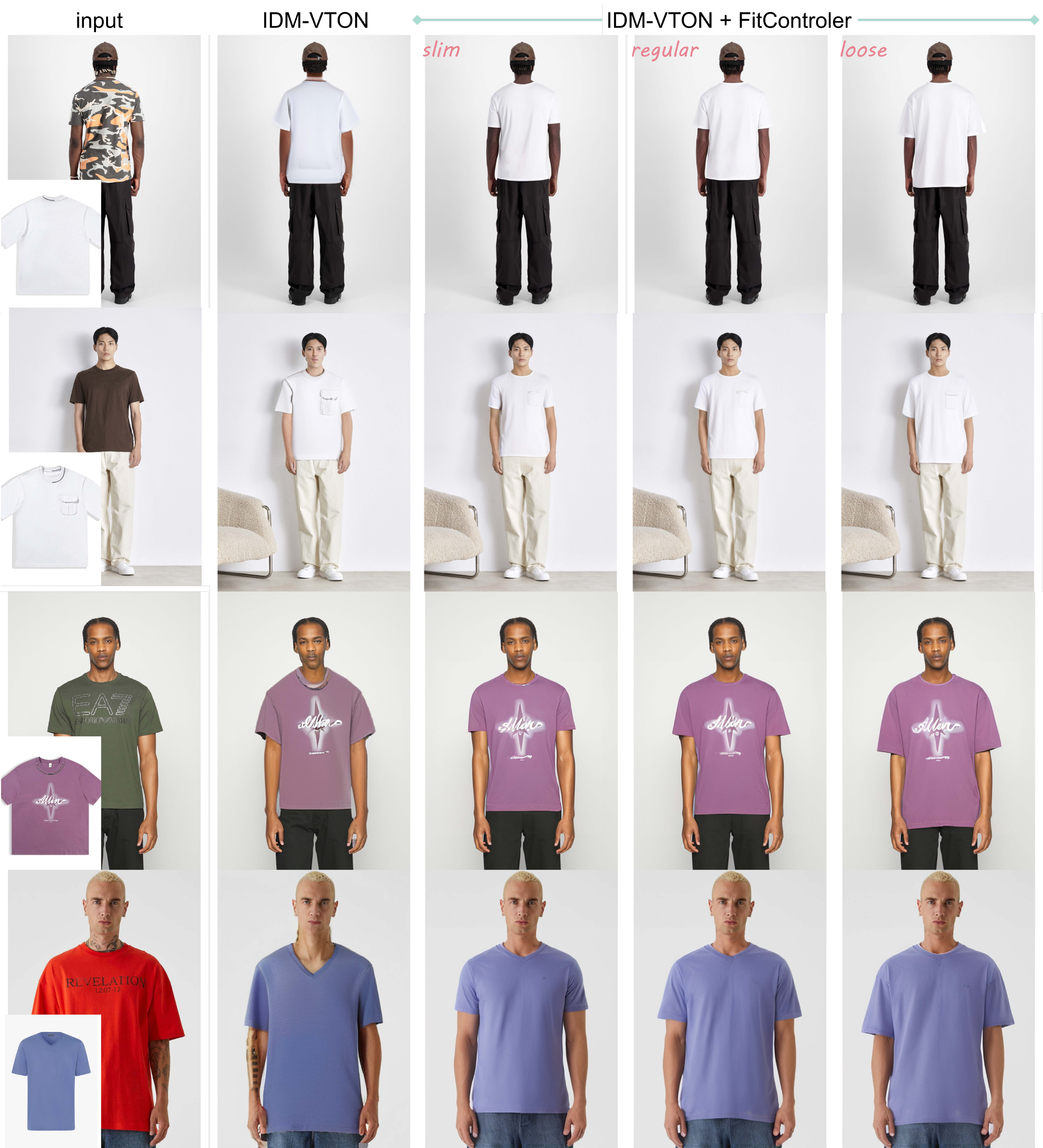}
   \caption{\textbf{Sample generated by IDM-VTON without and with FitControler on short-sleeve shirts.}
  FitControler provides reliable fit control and maintains high visual fidelity for a wide range of poses.}
   \label{fig:onecol4}
\end{figure*}

\begin{figure*}[t]
  \centering
   \includegraphics[width=\linewidth]{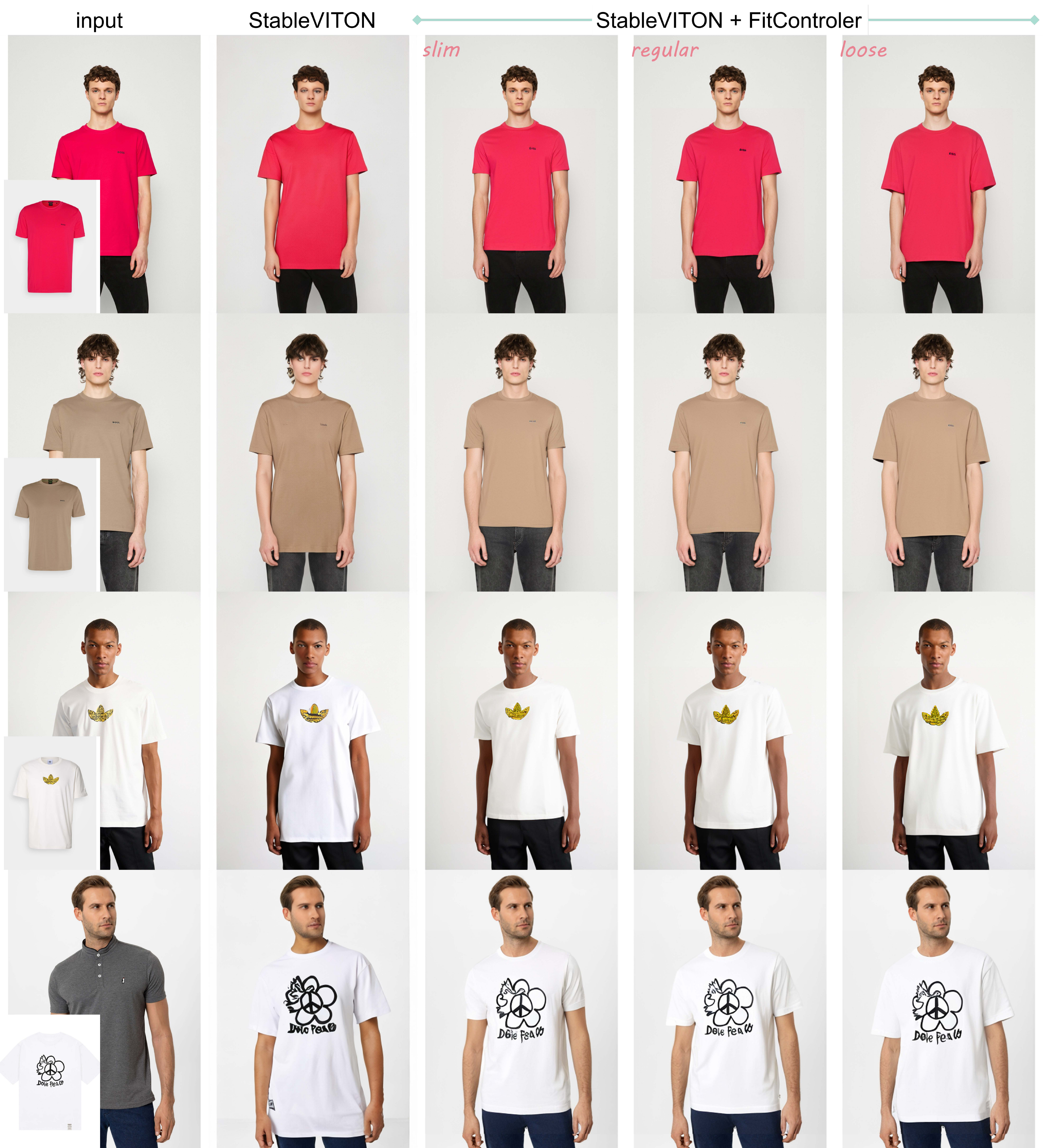}
   \caption{\textbf{Sample generated by StableVITON without and with FitControler on short-sleeve shirts.}
   FitControler provides reliable fit control and maintains high visual fidelity for a wide range of poses.}
   \label{fig:onecol5}
\end{figure*}

\begin{figure*}[t]
  \centering
   \includegraphics[width=0.8\linewidth]{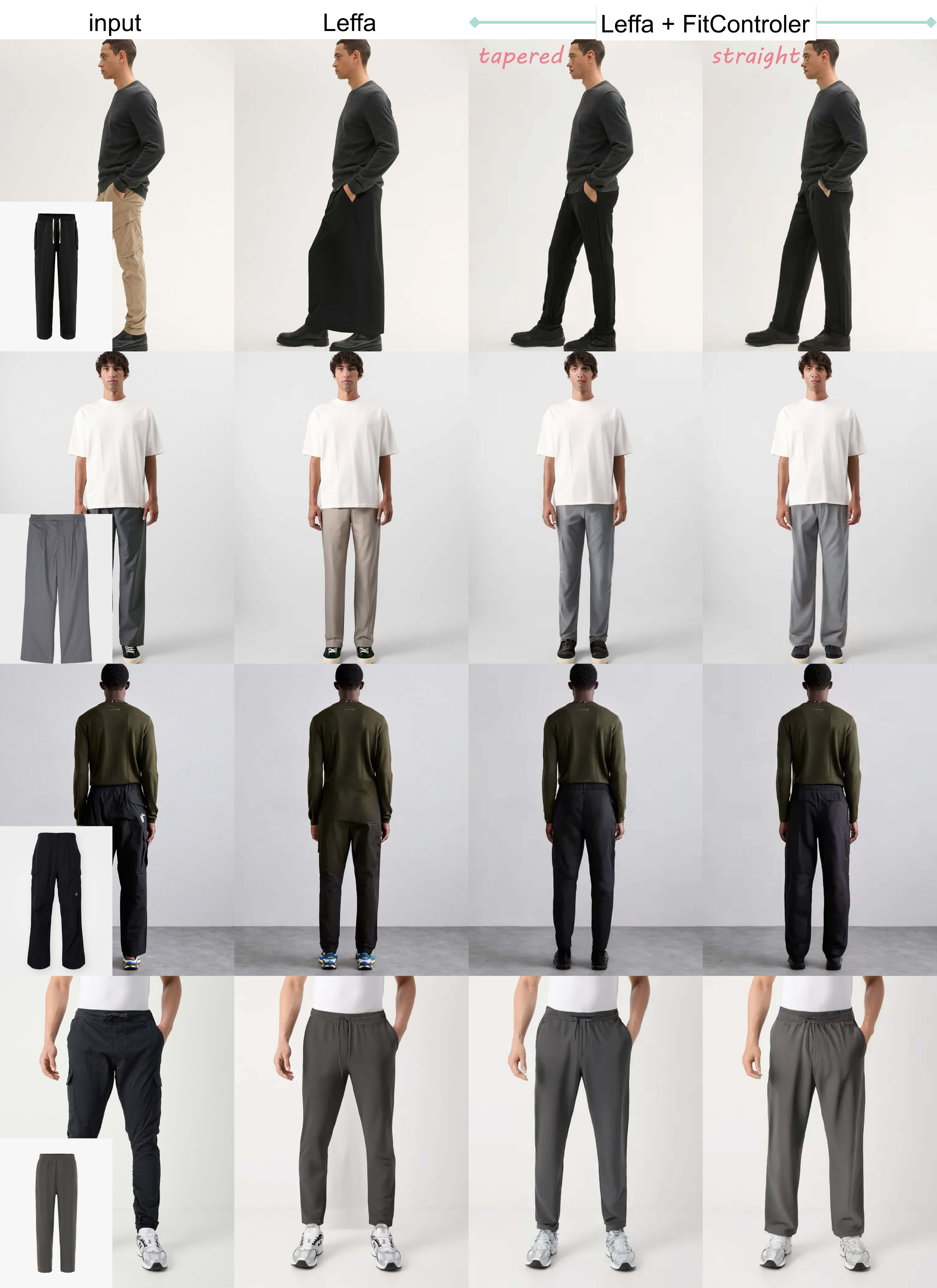}
   \caption{\textbf{Sample generated by Leffa without and with FitControler on trousers.}
   FitControler provides reliable fit control and maintains high visual fidelity for a wide range of poses.}
   \label{fig:onecol6}
\end{figure*}

\begin{figure*}[t]
  \centering
   \includegraphics[width=0.8\linewidth]{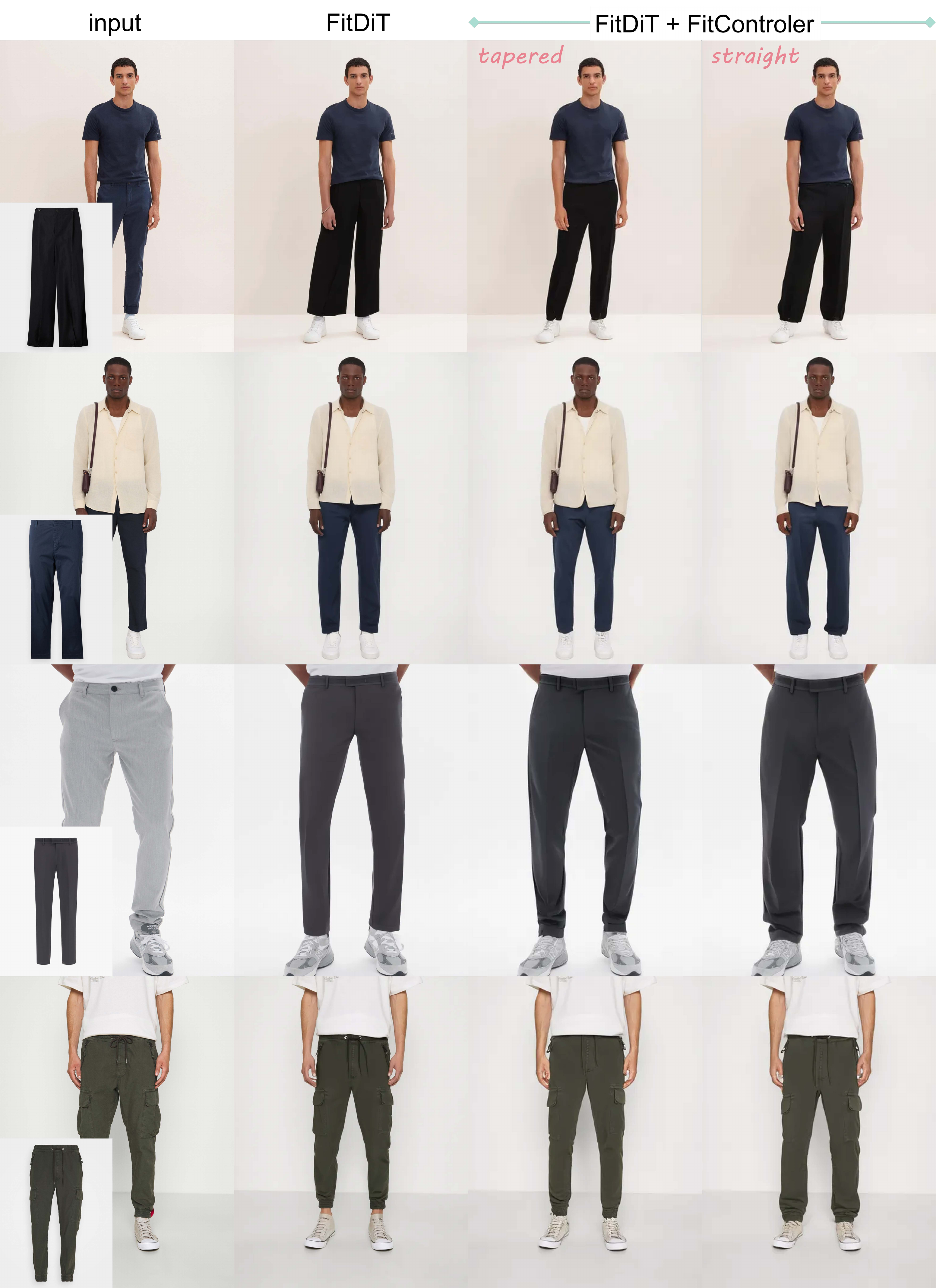}
   \caption{\textbf{Sample generated by FitDiT without and with FitControler on trousers.}
   FitControler provides reliable fit control and maintains high visual fidelity for a wide range of poses.}
   \label{fig:onecol7}
\end{figure*}

\begin{figure*}[t]
  \centering
   \includegraphics[width=0.8\linewidth]{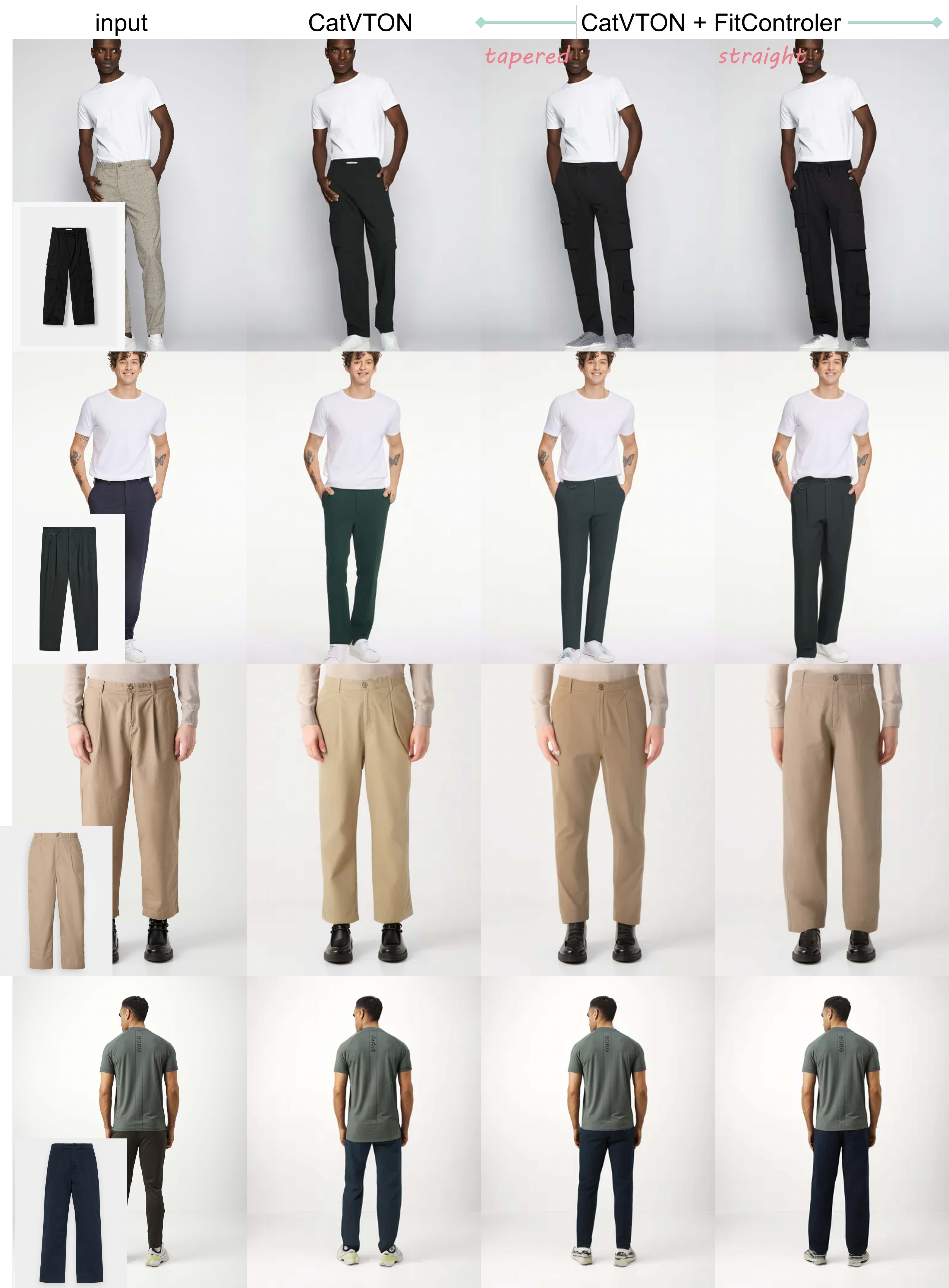}
   \caption{\textbf{Sample generated by CatVTON without and with FitControler on trousers.}
   FitControler provides reliable fit control and maintains high visual fidelity for a wide range of poses.}
   \label{fig:onecol8}
\end{figure*}

\begin{figure*}[t]
  \centering
   \includegraphics[width=0.8\linewidth]{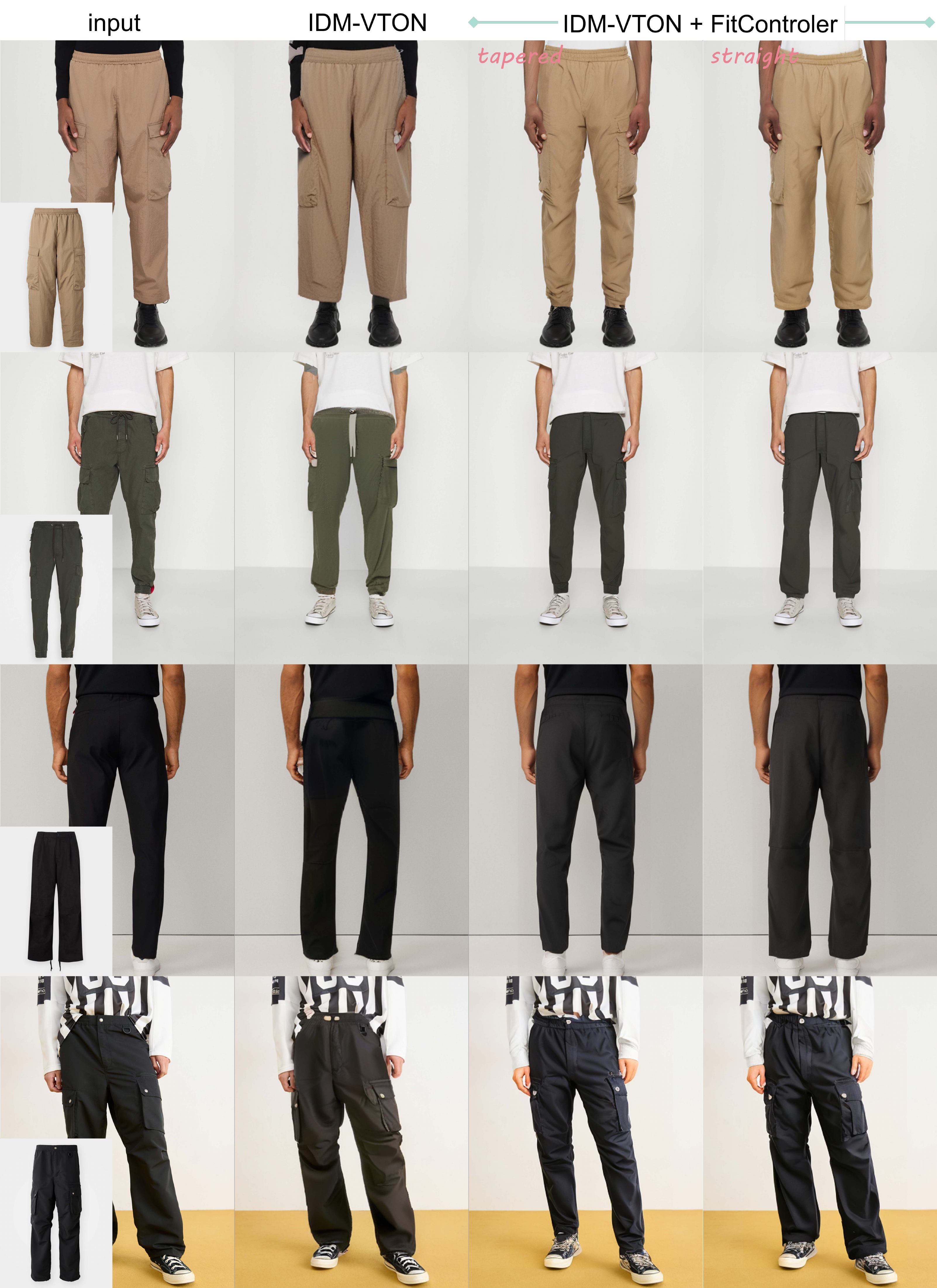}
   \caption{\textbf{Sample generated by IDM-VTON without and with FitControler on trousers.}
   FitControler provides reliable fit control and maintains high visual fidelity for a wide range of poses.}
   \label{fig:onecol9}
\end{figure*}

We further evaluate the generality of FitControler across multiple diffusion-based VTON frameworks, including CatVTON~\cite{chong2024catvton}, Leffa~\cite{zhou2025learning}, FitDiT~\cite{jiang2024fitdit}, IDM-VTON~\cite{choi2024improving}, and StableVITON~\cite{kim2024stableviton}. 
Figures~\ref{fig:onecol1}--\ref{fig:onecol5} present results on short-sleeve shirts, demonstrating precise and continuous control over garment tightness (\ie, \textit{slim}, \textit{regular}, and \textit{loose}). 
Figures~\ref{fig:onecol6}--\ref{fig:onecol9} show analogous results on trousers, further verifying that FitControler generalizes across garment categories while maintaining consistent fit-aware behavior across different backbone models.

\begin{figure}[t]
    \centering
    
    \begin{subfigure}[t]{0.25\textwidth}
        \centering
        \includegraphics[width=\textwidth]{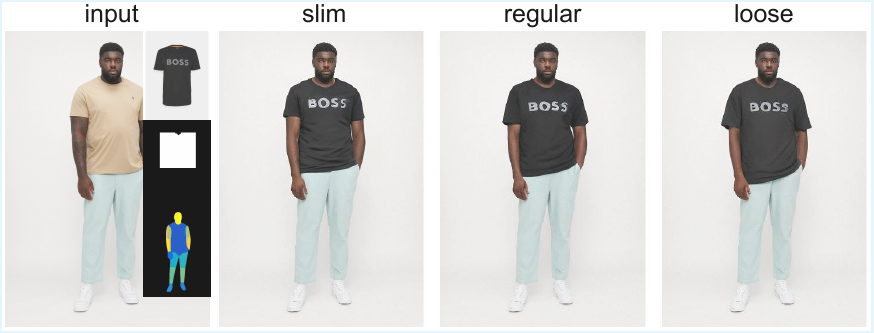}
        \caption{}
        \label{fig:left}
    \end{subfigure}%
    \hfill
    \begin{subfigure}[t]{0.74\textwidth}
        \centering
        \includegraphics[width=\textwidth]{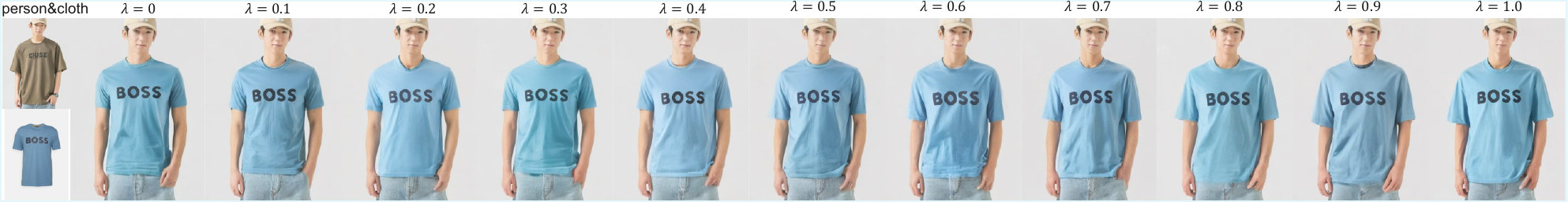} 
        \caption{}
        \label{fig:right}
    \end{subfigure}
\caption{\textbf{Limitations of FitControler.}
(a) Incomplete preservation of body shape on heavier body types.
(b) Visual discontinuities and artifacts during continuous fit control.}
    
\end{figure}
\section{Limitations and Future Work}
\label{sec:limitations}
Despite its effectiveness, FitControler has several limitations.

First, the dataset coverage remains limited. The current dataset primarily focuses on menswear and does not explicitly incorporate diverse body types. As illustrated in Fig.~\ref{fig:left}, although retaining the original garment-agnostic representation partially preserves the input physique, body shape consistency is not always maintained, and the overall visual quality may degrade in some cases. This limitation likely stems from insufficient exposure to diverse body shapes during training, which restricts generalization.

Second, our method is designed for discrete fit control. When extending it to continuous manipulation via feature interpolation,
\[
C_l = \lambda C^{\text{loose}}_l + (1-\lambda) C^{\text{slim}}_l,
\]
where $C_l$ denotes the output feature of the \textit{Fit Injector}, we observe color inconsistencies and control discontinuities. In particular, noticeable visual transitions may occur at certain interpolation ratios (e.g., around $\lambda=0.5$ and $\lambda=1$), suggesting that the learned fit representations are not smoothly organized in the feature space.

To address these limitations, future work will expand the dataset to include womenswear, additional garment categories with richer fit variations, and subjects with more diverse body shapes. Moreover, we plan to investigate continuous fit control strategies based on explicit geometric representations, such as garment keypoints, where interpolation in geometry space may enable smoother and more stable fit transitions.

\clearpage

\clearpage